\documentclass{article}

\newif\ifarxivversion
\arxivversiontrue
\newif\ifshowreproductioncode
\showreproductioncodetrue

\newcommand{\arxivauthors}{%
  \parbox[t]{0.45\textwidth}{\centering
    Kaito Takanami\\
    \normalfont\small
    Department of Physics, Graduate School of Science, The University of Tokyo\\
    Tokyo, Japan\\
    \texttt{takanami255@g.ecc.u-tokyo.ac.jp}}
  \And
  \parbox[t]{0.45\textwidth}{\centering
    Cengiz Pehlevan\\
    \normalfont\small
    John A. Paulson School of Engineering and Applied Sciences, Harvard University\\
    Kempner Institute for the Study of Natural and Artificial Intelligence, Harvard University\\
    Center for Brain Science, Harvard University\\
    Cambridge, MA, USA}}
\newcommand{\arxivacknowledgements}{%
  K.T. was supported by JST BOOST NAIS Grant Number JPMJBS2418.
  C.P. is supported by an NSF CAREER Award (IIS-2239780), DARPA grants DIAL-FP-038 and AIQ-HR00112520041, the Simons Collaboration on the Physics of Learning and Neural Computation, and the William F. Milton Fund from Harvard University.
  This work has been made possible in part by a gift from the Chan Zuckerberg Initiative Foundation to establish the Kempner Institute for the Study of Natural and Artificial Intelligence.
  The authors declare no competing interests.}

\usepackage{iclr2027_conference,times}
\usepackage[english]{babel}
\usepackage[T1]{fontenc}
\usepackage{microtype}
\usepackage{booktabs}
\usepackage{float}
\usepackage{titletoc}
\usepackage{xcolor}
\usepackage{url}
\usepackage{style/macros}

\numberwithin{equation}{section}

\title{How Does Chain-of-Thought Supervision Teach Models to Reason? A Theoretical Analysis
}

\ifarxivversion
  \author{\arxivauthors}
  \iclrfinalcopy
\else
  \author{Anonymous Authors}
\fi

\begin{document}

\maketitle
\ifarxivversion\lhead{Preprint}\fi

\begin{abstract}
  
Training with chain-of-thought (CoT) supervision is an effective approach for teaching models multi-step reasoning procedures. 
At inference time, however, models may need to continue reasoning beyond the finite teacher CoT sequences observed during training, and longer reasoning can instead degrade performance, a phenomenon known as overthinking. 
How the amount and structure of training-time CoT give rise to such failures remains theoretically poorly understood. 
To study this problem, we introduce an in-context weight prediction model trained on finite gradient descent trajectories for linear regression. 
Using high-dimensional asymptotic analysis, we show that the success and failure modes of long-horizon reasoning undergo phase transitions depending on the context length and the total amount of training CoT. 
We further distinguish two mechanisms of overthinking, error accumulation and error amplification, and show that with short training contexts, increasing the training CoT budget can worsen long-horizon error before eventually stabilizing reasoning. 
Experiments with fully trained linear and multi-head softmax attention further show that these qualitative behaviors persist.
Our results provide a theoretical account linking properties of training-time CoT to inference-time failures and offer guidance for designing teacher CoT that supports stable reasoning.

\end{abstract}

\section{Introduction}
\label{sec:introduction}

Popularized as a test-time prompting technique, chain-of-thought (CoT)~\citep{wei2022chain} is now widely used as training supervision to improve a model's reasoning ability~\citep{zelikman2022star,chung2024scaling}.
Whereas final-answer supervision provides only outcome-level feedback, supervising intermediate steps can help models learn general reasoning procedures.
Such CoT-based supervision has produced substantial gains on complex tasks such as mathematics and logical reasoning~\citep{ho2023reasoningteachers,zelikman2022star,lightman2023verify}.

During inference, a model may need to carry the reasoning procedure learned from CoT supervision beyond the number of steps shown during training to reach the correct answer.
In practice, extending reasoning can instead degrade performance, a phenomenon widely known as overthinking, so additional reasoning steps do not necessarily improve accuracy~\citep{bansal2022overthinking,su2025underthinkingoverthinkingempiricalstudy,hassid2025dont}.
Recent empirical studies suggest that overthinking is a problem involving the correctness and stability of the reasoning procedure itself.
For example, partial acquisition of reasoning capabilities may temporarily degrade long-horizon reasoning performance~\citep{wei2023inverse,wu2025ushaped,ren2026rethinking_sft_generalization}.
Moreover, systems with similar short-horizon performance may exhibit sharply different long-horizon failure modes: errors may simply accumulate across reasoning steps, or they may be amplified as the system becomes increasingly confident in and builds on its earlier mistakes~\citep{block2024butterfly,ainslie2026eigenanalysis}.
These findings show that locally plausible steps do not guarantee global convergence or error correction under repeated inference.

Despite these observations of long-horizon failures, it remains unclear how the amount and structure of training CoT shape the stability of reasoning.
In large language models, training data and model scale also affect performance, making the contribution of CoT supervision difficult to isolate experimentally~\citep{biderman2023pythia,li2024datacomp}.
Consequently, observed performance offers limited guidance on how to choose CoT training data that support reasoning.

To understand how training CoT affects reasoning theoretically, we introduce an in-context weight prediction model trained on gradient descent (GD) trajectories as teacher CoT.
Specifically, we use GD trajectories for linear-regression tasks as teacher CoT and train a linear attention model on each transition.
Previous work has shown that sufficient CoT supervision allows a model to approximately implement multi-step GD \citep{huang2025transformers}, but a theoretical understanding of long-horizon success conditions and failure regimes remains limited.
More recently, phase transitions between successful inference and overthinking have been identified in a setting without training-time CoT \citep{takanami2026asymptotic}.
We extend this framework to finite teacher CoT trajectories and ask whether recursive application of the learned model can reproduce the correct update rule beyond the training horizon.
Through this analysis, we characterize how the number of training tasks, context length, and teacher CoT length affect the inference dynamics.

Our contributions are as follows:
\begin{itemize}
  \item We derive a phase diagram in the asymptotic limit showing that context length and the total amount of training CoT divide long-horizon reasoning into four regimes: exact GD recovery, bounded error, exponential divergence, and quadratic divergence. (Section~\ref{sec:results})
  \item We show that reasoning can extend beyond the finite-length teacher CoT. The number of pretraining tasks required for this behavior decreases as the amount of teacher CoT increases, while shorter contexts require more teacher CoT. (Section~\ref{sec:recovering-gd})
  %\item We identify two distinct mechanisms of overthinking: errors accumulate when correction is missing, while unstable learned updates amplify existing errors. (Section~\ref{sec:two-mechanisms-overthinking})
  \item We identify two distinct mechanisms of overthinking: error accumulation, where errors introduced at each reasoning step remain uncorrected, and error amplification, where unstable reasoning dynamics repeatedly magnify existing errors, leading to quadratic and exponential MSE growth, respectively. (Section~\ref{sec:two-mechanisms-overthinking})
  \item We show that with short contexts, increasing the training CoT budget can initially degrade long-horizon performance before the model learns to reason correctly. (Section~\ref{sec:nonmonotonic-cot-budget})
  \item We empirically show that the same qualitative mechanisms persist in more complex architectures, including fully trained linear attention and multi-head softmax attention. (Section~\ref{sec:empirical-extensions})
\end{itemize}

Our results provide a unified theoretical understanding of the role of finite teacher CoT, appropriate CoT design as training data, and the mechanisms that generate failure.

\section{Related Work}

\paragraph{CoT supervision and overthinking.}
CoT data are increasingly used as training signals to improve reasoning and distill it into smaller models \citep{hsieh2023distilling,lightman2023verify,agarwal2024onpolicy,guo2025deepseek,zhu2025longreps,pengmei2025kinetics}.
At the same time, longer reasoning or recurrent computation can reduce performance, a phenomenon often described as overthinking~\citep{bansal2022overthinking,block2024butterfly,su2025underthinkingoverthinkingempiricalstudy,hassid2025dont,ainslie2026eigenanalysis}.
However, most existing work focuses on improving training methods or observing trained models, leaving unclear how the properties of training CoT determine test-time behavior.
We study this relation in a solvable model and characterize when training CoT supports correct inference and when it makes the test-time reasoning unstable.

\paragraph{Theoretical studies of CoT.}
Theoretical studies of CoT have mainly focused on its inference-time benefits, showing that CoT can enable efficient reasoning on otherwise difficult tasks \citep{abbe2024howfar,kim2025parity, huang2025transformers}.
However, the conditions under which such learning fails remain poorly understood.
To study this question theoretically, recent work extends analytically tractable models of in-context linear regression \citep{garg2022what,akyurek2023what,von2023transformers,zhang2025training,lu2025asymptotic} to iterative CoT, investigating how limited pretraining data or context and task mismatch can lead to overthinking \citep{javanmard2026understanding,takanami2026asymptotic}.
However, these studies mainly analyze the test-time behavior of trained models and do not characterize the effect of providing finite teacher CoT as training data.
We therefore focus on training-time CoT and ask how far a model can continue to reason correctly beyond the reasoning steps observed during training and how this generalization fails.

\section{Model}
\label{sec:model}

We study an in-context weight-prediction model for linear regression.
In this model, training CoT consists of GD trajectories, and the pre-trained model is recursively applied as test-time CoT to reproduce GD in context on new tasks.

\subsection{Data generation}
We consider \(M\) linear-regression tasks indexed by \(\mu=1,\ldots,M\).
For each task \(\mu\), we prepare the training data and teacher CoT trajectories as follows.

\paragraph{Training tasks.}
For each of the $M$ tasks, we prepare $L$ labeled examples ($\{\vb{x}_{\mu, l}, y_{\mu,l}\}_{l=1,\cdots, L}$), given by
\begin{align}
    \mat X_\mu :=
    [\vb x_{\mu,1},\ldots,\vb x_{\mu,L}]
    \in\R^{D\times L},
    \quad 
    \vb y_\mu
    :=
    \mat X_\mu^\top\vb w_\mu+\vb\varepsilon_\mu \in \R^{L},
    \label{eq:training-task}
\end{align}
where $\vb x_{\mu,l}\sim\mathcal N(\vb0_D,\mat I_D)$, $\vb w_\mu\sim\mathcal N\ab(\vb0_D,\mat I_D/D\ab)$,
and  $\vb\varepsilon_\mu\sim\mathcal N(\vb0_L,\sigma^2\mat I_L)$, independently across all tasks and examples.
We denote the task context by
\(Q_\mu:=(\mat X_\mu,\vb y_\mu)\).

\paragraph{CoT as training data.}
We prepare a \(K\)-step trajectory of gradient descent (GD) $\{\vb{s}_{\mu, k}\}_{k=0,\cdots, K}$ with step size \(\eta>0\) on the empirical squared loss.
Starting from zero, its states satisfy
\begin{align}
    \vb s_{\mu,k+1}
    &:={}
    \vb s_{\mu,k}
    -\frac{\eta}{L}\mat X_\mu
      \ab(\mat X_\mu^\top\vb s_{\mu,k}-\vb y_\mu\ab)
    \qquad (k=0,\ldots,K-1),
    \label{eq:teacher-gd-trajectory}
\end{align}
and $\vb s_{\mu,0}:={}\vb0_D$.

\subsection{Pre-training setting}
During pre-training, the model is trained to predict the next step of the teacher CoT, $\vb{s}_{\mu,k+1}$, from the context and the teacher CoT at step $k$, i.e., $(Q_\mu,\vb{s}_{\mu,k})$. 
Specifically, the input data (prompt), architecture, and loss function are defined as follows.

\paragraph{Prompt construction.}
At training step \(k\), the prompt contains the original \(L\) labeled examples and a slot for the current teacher state:
\begin{align}
    \mat E_{\mu,k}
    &:={}
    \begin{pmatrix}
      \mat X_\mu&\vb0_D\\
      \vb y_\mu^\top&0\\
      \mat0_{D\times L}&\vb s_{\mu,k}\\
      \vb0_L^\top&1
    \end{pmatrix}
    =
    \begin{pmatrix}
      \vb x_{\mu,1}&\cdots&\vb x_{\mu,L}&\vb0_D\\
      y_{\mu,1}&\cdots&y_{\mu,L}&0\\
      \vb0_D&\cdots&\vb0_D&\vb s_{\mu,k}\\
      0&\cdots&0&1
    \end{pmatrix}
    \in\R^{(2D+2)\times(L+1)}.
    \label{eq:step-prompt}
\end{align}
The first \(L\) columns encode labeled examples.
The final column is a state slot, and the marker $1$ in the last row distinguishes it from an example token.
The prompt for predicting \(\mathbf{s}_{\mu,k+1}\) contains the task context \(Q_\mu\) and the current state \(\mathbf{s}_{\mu,k}\), but no earlier states. 
Thus, conditional on the task context, the one-step predictor is Markovian.
Masking or downweighting earlier reasoning tokens in this way can improve length generalization~\citep{abbe2024howfar,liu2023transformers}.

\paragraph{Linear attention.}
For a prompt with $L$ context examples, we use a one-layer linear self-attention model with a residual connection, defined by
\begin{align}
    \mathsf{Atten}(\mat E; \theta)
    &:={}
    \mat E
    +\frac{1}{L}
      (\mat W_V\mat E)
      (\mat W_K\mat E)^\top
      \mat W_Q\mat E.
    \label{eq:linear-attention-operator}
\end{align}
Here the trainable parameters are \(\theta=(\mat W_Q,\mat W_K,\mat W_V)\), where \(\mat W_Q,\mat W_K,\mat W_V\in\R^{(2D+2)\times(2D+2)}\) are the query, key, and value matrices, respectively.
We read the state slot of the final column using
\begin{align}
    \mathsf{Readout}(\mat E)
    &:=
    \mat E_{D+2:2D+1,\,\mathrm{end}}.
    \label{eq:state-readout-operator}
\end{align}
The next-step predictor of the weight vector is therefore
\begin{align}
    \hat{\vb{s}}_{\mu,k+1}
    &:={}
    \mathsf{Readout}\ab(
      \mathsf{Atten}\ab(
        \mat E_{\mu,k}; \theta
      \ab)
    \ab).
    \label{eq:full-one-step-predictor}
\end{align}

\paragraph{Learning.}
We consider the squared one-step prediction error over all teacher transitions in pre-training data:
\begin{align}
    \mathcal L(\theta)
    &:={}
    \frac{1}{MK}\sum_{\mu=1}^{M}\sum_{k=0}^{K-1}
    \ab\|\hat{\vb s}_{\mu,k+1}-\vb s_{\mu,k+1}\ab\|_2^2.
    \label{eq:one-step-training-loss}
\end{align}

\subsection{Effective model}
\label{sec:effective-model}
For analytical simplicity, we impose the following constraints on the trainable parameters:
\begin{align}
    \mat W_V
    :={}
    \begin{pmatrix}
      \ast&\ast&\ast&\ast\\
      \ast&\ast&\ast&\ast\\
      \mat I_D&\vb0_D&\mat I_D&\vb0_D\\
      \ast&\ast&\ast&\ast
    \end{pmatrix},\quad
    \mat W_K^\top\mat W_Q
    :={}
    \begin{pmatrix}
      \ast&\ast&\mat A&\vb0_D\\
      \ast&\ast&\vb0_D^\top&\eta \\
      \ast&\ast&\mat0_{D\times D}&\vb0_D\\
      \ast&\ast&\vb0_D^\top&0
    \end{pmatrix},
    \label{eq:effective-attention-parameterization}
\end{align}
where the rows from top to bottom and the columns from left to right are both partitioned into blocks of dimensions \((D,1,D,1)\).
Here \(\ast\) denotes a block that does not affect the readout, while \(\mat A\in\R^{D\times D}\) is the only learnable parameter.
%This restriction is motivated by the rotational symmetry of the data and by related tractable attention models \citep{huang2025transformers,javanmard2026understanding,takanami2026asymptotic}.
This restriction is less severe than it may appear: each constrained block is justified by one of three considerations:
(i) the rotational symmetry of the data, which motivates choosing the fixed blocks so that they do not select preferred directions;
(ii) the fact that fixing it entails no loss of generality in the setting studied here; or
(iii) preventing the model from advancing its inference without consulting the context
(see Appendix~\ref{app:effective-parameterization} for details).
These reductions concentrate the nontrivial parameter governing inference in the single matrix \(\mat A\).
Moreover, gradient-flow analyses of related models show that training from suitable initializations leads to analogous block structures~\citep{huang2025transformers,javanmard2026understanding}.
%In Section~\ref{sec:empirical-extensions}, we further show that the same qualitative behavior persists when the attention parameters are trained end-to-end without fixing these blocks.
Under this parameterization, Equation~\eqref{eq:full-one-step-predictor} is given by
\begin{align}
    \hat{\vb{s}}_{\mu, k+1}
    &={}
    \vb s_{\mu, k}+\frac{1}{L}\mat X_\mu
    \ab(\mat X_\mu^\top\mat A\vb s_{\mu, k}+\eta\vb y_\mu\ab),
    \label{eq:effective-update}
\end{align}
and the effective learning problem derived from Equation~\eqref{eq:one-step-training-loss} is  
\begin{align}
    \mathcal L_{\mathrm{eff}}(\mat A)
    :={}
    \frac{1}{MK}
    \sum_{\mu=1}^M\sum_{k=0}^{K-1}
    \ab\|\hat{\vb{s}}_{\mu, k+1}
      -\vb s_{\mu,k+1}\ab\|_2^2
    +\frac{\lambda}{D}\ab\|\mat A\ab\|_{\mathrm F}^2,
    \label{eq:effective-regularized-loss}
\end{align}
where we introduced the regularization parameter $\lambda>0$.
For \(\lambda>0\), the objective is strongly convex, so
\(\hat{\mat A}:=\argmin_{\mat A}\mathcal L_{\mathrm{eff}}(\mat A)\) is unique.

\subsection{Inference setting}
At inference time, we model test-time CoT by fixing the parameters of the trained model and applying the model recursively.
Despite its simplicity, this recurrent form of CoT has recently regained attention because it enables flexible reasoning without relying on language tokens \citep{hao2025training,geiping2025scaling,shen2025codi,zhu2025reasoning}.

\paragraph{Test-time CoT.}
Consider an independent task \(Q^{\mathrm{in}}=(\mat X^{\mathrm{in}},
\vb y^{\mathrm{in}})\), drawn from the same distribution as the training tasks.
Explicitly, its examples and
labels are generated by
\begin{align}
    \mat X^{\mathrm{in}}
    &:=
    [\vb x^{\mathrm{in}}_1,\ldots,\vb x^{\mathrm{in}}_{L}]
    \in\R^{D\times L},
    &
    \vb y^{\mathrm{in}}
    &:=
    (\mat X^{\mathrm{in}})^\top\vb w^{\mathrm{in}}+\vb\varepsilon^{\mathrm{in}}
    \in\R^{L},
\end{align}
where $\vb x^{\mathrm{in}}_l\sim\mathcal N(\vb0_D,\mat I_D)$, $\vb w^{\mathrm{in}}\sim\mathcal N\ab(\vb0_D,\mat I_D/D\ab)$, and $\vb\varepsilon^{\mathrm{in}}\sim\mathcal N(\vb0_{L},\sigma^2\mat I_{L})$.
At inference step \(t\), the prompt is constructed in the same way as in Equation~\eqref{eq:step-prompt}:
\begin{align}
  P_t
  &:={}
    \begin{pmatrix}
      \mat X^{\mathrm{in}}&\vb0_D\\
      (\vb y^{\mathrm{in}})^\top&0\\
      \mat0_{D\times L}&\hat{\vb w}_{t}\\
      \vb0_{L}^\top&1
    \end{pmatrix}
  =
    \begin{pmatrix}
      \vb x^{\mathrm{in}}_1&\cdots&\vb x^{\mathrm{in}}_{L}&\vb0_D\\
      y^{\mathrm{in}}_1&\cdots&y^{\mathrm{in}}_{L}&0\\
      \vb0_D&\cdots&\vb0_D&\hat{\vb w}_t\\
      0&\cdots&0&1
    \end{pmatrix}
    \in\R^{(2D+2)\times(L+1)},
\end{align}
where \(\hat{\vb w}_t\) is the current estimate of \(\vb w^{\mathrm{in}}\), and \(\hat{\vb w}_0=\vb0_D\).
Then the test-time CoT recursion is
\begin{align}
  \hat{\vb w}_{t+1}
  = \mathsf{Readout}\ab(\mathsf{Atten}(P_t; \hat{\mat{A}}))
  =\hat{\vb w}_t
   +\frac{1}{L}\mat X^{\mathrm{in}}
     \left[(\mat X^{\mathrm{in}})^\top\hat{\mat A}
       \hat{\vb w}_t+\eta\vb y^{\mathrm{in}}\right].
  \label{eq:learned-iteration}
\end{align}

\paragraph{Evaluation.}
We evaluate the estimate at inference time $t$ by its mean squared error:
\begin{align}
    \mathcal E(t)
    &:={}
    \E\ab\|\hat{\vb w}_{t}-\vb w^{\mathrm{in}}\ab\|_2^2,
    \label{eq:finite-risk}
\end{align}
where the expectation is over the training and test data.
For comparison, we define the generalization error when GD is successfully reproduced as
\begin{align}
    \mathcal E_{\mathrm{GD}}(t)
    &:={}
    \E\ab\|\vb s_t^{\mathrm{GD}}-\vb w^{\mathrm{in}}\ab\|_2^2,
\end{align}
where $\vb s_t^{\mathrm{GD}}$ denotes the GD trajectory at step $t$ on the test context, with initial condition $\vb s_0^{\mathrm{GD}} = \vb{0}_D$.
When the model perfectly learns to generate the teacher-provided CoT, we have $\mathcal{E}(t) = \mathcal{E}_{\mathrm{GD}}(t)$.

\subsection{High-dimensional limit}
To obtain a nontrivial regime in which contextual information and training-task diversity remain comparable to the problem dimension, we take the proportional limit $D,L,M\longrightarrow\infty$ with
\begin{align}
    \frac{L}{D}\longrightarrow\alpha,
    \quad
    \frac{M}{D}\longrightarrow\tau,
    \qquad K,t=O(1).
    \label{eq:proportional-limit}
\end{align}
Here, \(\alpha\) and \(\tau\) measure the numbers of context examples and training tasks per dimension, respectively, while the training and test-time CoT depths remain finite.
We are particularly interested in $t>K$, where the model must continue inference beyond the GD steps observed in the teacher CoT.

\section{Precise Characterization of Test-Time CoT Dynamics}
\label{sec:results}

The central question is how the generalization error \(\mathcal E(t)\) evolves with test-time depth: when does the CoT inference reproduce GD, and what form does its failure take otherwise?
To this end, we first seek a precise characterization of \(\mathcal E(t)\).
We introduce \textit{training CoT budget} as 
\begin{align}
  \varphi:=\tau(K-1).
\end{align}
Since it measures the total number of CoT transitions in the training data, \(\varphi\) can be interpreted as the computational budget used to prepare the training CoT trajectories.

In the limit $\lambda\to0$, the CoT inference introduced in Equation~\eqref{eq:learned-iteration} gives
\begin{equation}
  \hat{\vb w}_t=\mat Q_t\mat X^{\mathrm{in}}\vb y^{\mathrm{in}} / L,
  \label{eq:main-test-response}
\end{equation}
where we introduced
\begin{align}
  \mat H:=\frac{1}{L}\mat X^{\mathrm{in}}
      (\mat X^{\mathrm{in}})^\top,
  \quad
  \mat Q_t:=\eta\sum_{j=0}^{t-1}
      (\mat I_D+\mat H\hat{\mat A})^j.
  \label{eq:inference-filter}
\end{align}
Consequently, the generalization error in Equation~\eqref{eq:finite-risk} is given by
\begin{equation}
  \mathcal E(t)
  =1-\frac2D\E\operatorname{tr}(\mat Q_t\mat H)
   +\frac1D\E\|\mat Q_t\mat H\|_{\mathrm F}^2
   +\frac{\sigma^2}{L}
      \E\operatorname{tr}(\mat Q_t\mat H\mat Q_t^\top).
  \label{eq:main-risk-reduction}
\end{equation}
The last two terms in Equation~\eqref{eq:main-risk-reduction} mix
\(\mat Q_t\) and \(\mat Q_t^\top\).
Because \(\mat Q_t\) is a polynomial in \(\mat H\hat{\mat A}\), evaluating
these terms at arbitrary depth $t$ requires not only its eigenvalue distribution
but the full family of mixed \(*\)-moments involving \(\mat H\hat{\mat A}\) and \(\hat{\mat A}^\top\mat H\).
To overcome this difficulty, \citet{takanami2026asymptotic} introduced two-point generating functions and computed them by iteratively applying the cavity method in random matrix theory.
In the present setting, the training CoT trajectories couple successive transitions, so the resulting cavity equations do not close under the same recursion.
However, the two-point problem reduces to one-point Marchenko-Pastur averages when the learned matrix takes the form \(\hat{\mat A}=-\eta\mat P\) for an orthogonal projector \(\mat P\).
In this case, the learned update recovers full GD or performs GD only on the subspace learned during pretraining.
We refer to the latter as projected GD.
This reduction yields the following result.

\begin{result}[Finite-depth error in GD or projected-GD regimes]
\label{res:finite-depth-rich-context}
Consider \(\lambda\to0\).
For \(\rho>0\), let \(\nu_\rho\) denote the Marchenko--Pastur law with sample-to-dimension ratio \(\rho\), and define
\begin{equation}
  \langle f\rangle_\rho:=\int f(x)\,\mathrm d\nu_\rho(x),
  \quad
  q_t(x):=\eta\sum_{j=0}^{t-1}(1-\eta x)^j,
  \quad
  \kappa_{j,t}:=(\alpha-\varphi)(\eta t)^j
      +\varphi\langle q_t^j\rangle_{\alpha/\varphi}.
  \label{eq:mp-average-gd-filter}
\end{equation}

If \(\varphi>\max\{1,1/\alpha\}\), the learned update is GD and
\begin{equation}
  \mathcal E(t)=\mathcal E_{\mathrm{GD}}(t)
  =\langle(1-xq_t)^2\rangle_\alpha
   +\frac{\sigma^2}{\alpha}\langle xq_t^2\rangle_\alpha.
  \label{eq:main-recovered-gd-risk}
\end{equation}
If \(\alpha>1\) and \(0<\varphi<1\), the learned update is projected GD and
\begin{align}
  \mathcal E(t)
  =&1-\varphi+\varphi\langle(1-xq_t)^2\rangle_{\alpha/\varphi}
  +\frac{\varphi[2(1-\varphi)+\sigma^2]}{\alpha}
       \langle xq_t^2\rangle_{\alpha/\varphi}\notag\\
  &-\frac{2(1-\varphi)}{\alpha}\kappa_{1,t}
  +\frac{(1-\varphi)(1-\varphi+\sigma^2)}{\alpha^2}\kappa_{2,t}
  +\frac{1-\varphi}{\alpha^2}\kappa_{1,t}^2.
  \label{eq:phase-two-gd-risk}
\end{align}
\end{result}
This result characterizes the regimes in which the learned update recovers full GD or projected GD, enabling a complete analytical characterization of its test-time dynamics.
In the remaining regime, evaluating Equation~\eqref{eq:main-risk-reduction} at finite $t$ still requires the two-point functions.
Instead, to isolate the failure mechanisms of interest, we study \(\mathcal E(t)\) as \(t\to\infty\).
As Equation~\eqref{eq:inference-filter} shows, \(\mat Q_t\) is a sum of powers of the inference matrix.
In this limit, the relevant dominant eigenvalues alone determine whether \(\mathcal E(t)\) remains bounded, grows polynomially, or diverges exponentially.
The required one-point spectral quantities can be obtained from the cavity method in random matrix theory.
Combining them with the GD and projected GD regimes above yields the following four phases.

\begin{figure}[t]
  \centering
  \includegraphics[width=1.0\linewidth]{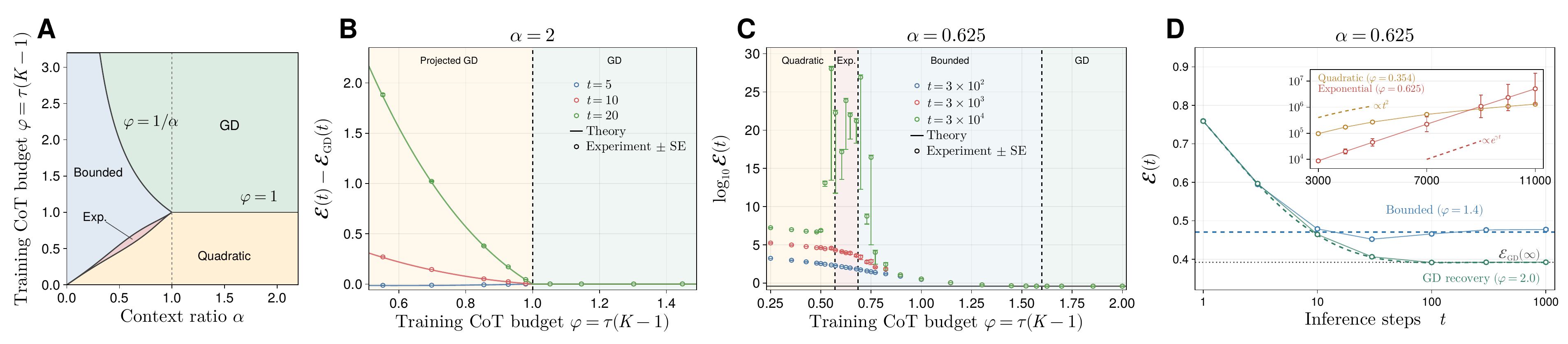}
  \caption{
    \textbf{Training CoT budget and context length control GD recovery and test-time error growth.}
    (A) Theoretical phase diagram in the context ratio $\alpha$ and training CoT budget $\varphi=\tau(K-1)$.
    The four regions correspond to GD recovery, bounded error, exponential
    divergence, and quadratic divergence.
    (B) Numerical experiments (circles) and theoretical prediction (solid line) on excess generalization error $\mathcal E(t)-\mathcal E_{\mathrm{GD}}(t)$ at $\alpha=2$.
    (C) Numerical experiments of the generalization error at $\alpha=0.625$.
    (D) Typical test-time error dynamics in the four phases;
    the dotted lines show the theoretical long-time predictions, and the inset highlights the quadratic and exponential regimes at large $t$.
    (B,C) The vertical dotted lines show the theoretical phase boundaries.
    Parameters: (B,C) $D=96$, $K=2$, $\eta=0.15$ and $\sigma^2=0.01$.
    (B-D) Error bars represent one standard error across eight independent runs.
  }
  \label{fig:training-cot-budget-phases}
\end{figure}

\begin{result}[Long-time test-time CoT dynamics]
\label{res:long-time-phases}
There exist critical values \(\varphi_-(\alpha)\) and \(\varphi_+(\alpha)\) such that, for sufficiently small \(\eta>0\) and as \(t\to\infty\),
\begin{equation}
\mathcal E(t)=
\begin{cases}
\mathcal E_{\mathrm{GD}}(t),
& \varphi>\max\{1,1/\alpha\}\\
O(1),
& 0<\alpha<1,\quad \varphi_+<\varphi<1/\alpha\\
\exp\ab(\Theta(\eta t)),
& 0<\alpha<1,\quad \varphi_-<\varphi<\varphi_+\\
\Theta(t^2),
& (0<\alpha<1,\ 0<\varphi<\varphi_-)
             \textnormal{ or }(\alpha>1,\ 0<\varphi<1) .
\end{cases}
\label{eq:main-long-time-phases}
\end{equation}
\end{result}

Appendix~\ref{app:phase-summary} lists the corresponding derivations.
Figure~\ref{fig:training-cot-budget-phases} summarizes the phase diagram and its agreement with numerical experiments.

Taken together, Results~\ref{res:finite-depth-rich-context} and \ref{res:long-time-phases}  show that the learned update and its long-time behavior change qualitatively with the training CoT budget and the context length.
With sufficient budget, the learned update recovers GD, whereas below this threshold the error can remain bounded or diverge polynomially or exponentially.
The next section discusses the implications of this phase diagram.

\section{Insights from the Test-Time CoT Phase Diagram}
\label{sec:discussion}

In the previous section, we theoretically derived the test-time dynamics of the generalization error in the regimes where the learned update recovers GD or projected GD, and identified the phase transitions governing the long-time inference limit.
In this section, we discuss what these results reveal about the effects of training CoT, their practical implications, and related empirical observations.

\subsection{Recovering GD from Finite-Length CoT}
\label{sec:recovering-gd}

In this subsection, we discuss the GD recovery phase and the conditions that determine its boundary.

\paragraph{Generalization beyond the training horizon.}
The first insight is that finite-length teacher CoT can teach the full GD update rule.
Although training trajectories contain only \(K\) steps, Result~\ref{res:finite-depth-rich-context} shows that, in the recovery phase, the test-time recursion matches GD beyond the training horizon.
From a practical perspective, increasing the training CoT length \(K\) reduces the number of tasks needed to recover GD.
This provides a practical trade-off in training-data design: when collecting diverse tasks and in-context examples is costly, longer teacher CoT can be used instead.

\paragraph{CoT budget matters more than task diversity or CoT length alone.}
In our model, Results~\ref{res:finite-depth-rich-context} and~\ref{res:long-time-phases} show that the recovery condition depends on the number of training tasks and the training CoT length \(K\) only through the CoT budget \(\varphi=\tau(K-1)\).
To understand this mechanism intuitively, consider \(\alpha>1\): there are \(D^2\) parameters to learn and \(M(K-1)\) teacher steps, each providing \(D\) scalar constraints.
This gives the recovery condition
\begin{equation}
  M(K-1)D>D^2
  \quad\Longleftrightarrow\quad
  \varphi = M(K-1) / D>1.
\end{equation}
Because each teacher step adds a linearly independent state direction until the accumulated state matrix has rank \(D\), the recovery condition depends on \(\tau\) and \(K\) only through the CoT budget \(\varphi\).

More generally, our result suggests that the value of teacher CoT is determined not by the number of steps, but by the diversity of states used to constrain the update rule.
If successive teacher states are redundant, shorter teacher CoT from more diverse tasks may be more informative. 
Conversely, a longer teacher CoT may compensate for fewer tasks when it contains sufficiently diverse states.

\paragraph{Context length determines the CoT budget required.}
For \(0<\alpha<1\), however, the CoT budget required for recovery depends on the context length ($\varphi > 1/\alpha$).
This dependence can be understood through a simple analogy.
Suppose the task is sorting, but the context contains only examples of lists without duplicates, such as \([2,1,3]\).
In this case, sorting and removing duplicates before sorting produce the same output $[1,2,3]$, so matching the examples does not uniquely identify the intended rule.
However, an example containing duplicates, such as \([1,2,2,3]\), distinguishes the two rules: sorting returns \([1,2,2,3]\), whereas removing duplicates before sorting returns \([1,2,3]\).
Similarly, in our setting, a short context constrains the update rule only on a subspace, so a larger teacher CoT budget across contexts is required to identify it.
In our model, this trade-off is reflected in the recovery condition \(\alpha\varphi>1\) in Result~\ref{res:long-time-phases}.
Practically, this suggests that the teacher CoT budget should not be chosen independently of the context length.

\subsection{Two Mechanisms of Overthinking}
\label{sec:two-mechanisms-overthinking}

Figure~\ref{fig:training-cot-budget-phases}A shows that, in the divergent regimes, test-time error grows either polynomially or exponentially, depending on the training CoT budget and context length.
To explain these behaviors, we formulate inference as a state transition and characterize polynomial overthinking as \textit{error accumulation} and exponential overthinking as \emph{error amplification}~\citep{ainslie2026eigenanalysis}.
The test-time dynamics in Equation~\eqref{eq:learned-iteration} can be written as
\begin{align}
  \hat{\vb{w}}_{t+1}=\mat{J}\hat{\vb{w}}_t+ (\eta / L) \mat{X}^{\mathrm{in}}\vb{y}^{\mathrm{in}},
  \label{eq:Jacobian}
\end{align}
where $\mat{J}=\mat{I}+\mat{H}\hat{\mat{A}}$.
Here, the first term propagates the current state, while the second term injects input from the context at every step.
Error evolution is determined by the eigenvalue structure of \(\mat{J}\).

%For error accumulation, suppose that \(\mat{J}\) has an eigenmode with eigenvalue \(1\).
%Let \({z}_t\) and \(c\neq0\) denote the state and input components along this mode, respectively.
%They satisfy
%\begin{align}
%  z_{t+1}=z_t+\eta c,
%\end{align}
%which gives $z_t=\eta ct$ for $z_0=0$.
%This represents a situation in which the model incorporates incoming information but lacks sufficient self-correction based on the information already incorporated.
%Because the input at each step accumulates without being canceled, the estimate grows in proportion to \(t\), and MSE diverges polynomially as \(t^2\).For error accumulation, consider the polynomial regime in Result~\ref{res:long-time-phases}.

For error accumulation, consider the polynomial regime in Result~\ref{res:long-time-phases}.
In this regime, we show that \(\mat J\) has an eigenmode with eigenvalue \(1\) that is excited by the context input (see Appendices~\ref{app:proof-phase-two} and~\ref{app:proof-phase-three-long-time}).
Let \(z_t\) and \(c\neq0\) denote the state and input components along one such excited mode, respectively.
Projecting Equation~\eqref{eq:Jacobian} onto this mode gives
\begin{align}
  z_{t+1}=z_t+\eta c.
\end{align}
Iterating this recursion from \(z_0=0\) yields \(z_t=\eta ct\).
Here, the unit coefficient on \(z_t\) means that the previously accumulated state persists unchanged from one step to the next, so the input added at each step accumulates rather than being canceled.
Consequently, the estimate grows in proportion to \(t\), and the MSE diverges as \(t^2\).
In the projected-GD regime (\(\alpha>1\) and \(\varphi<1\)), these unit-eigenvalue modes correspond to directions outside the subspace learned during pretraining.

For error amplification, consider the exponential regime in Result~\ref{res:long-time-phases}.
In this regime, we show that \(\mat J\) has an unstable eigenmode with eigenvalue \(r\) satisfying \(|r|>1\), and the context input excites this mode (see Proposition~\ref{prop:long-time-cavity-phases}).
Let \(z_t\) and \(c\neq0\) denote, respectively, the state and input components associated with this mode. Projecting Equation~\eqref{eq:Jacobian} onto the mode gives
\begin{align}
  z_{t+1}=rz_t+\eta c.
\end{align}
Iterating this recursion from \(z_0=0\) yields \(z_t=\eta c\,(r^t-1)/(r-1)\), which grows exponentially with \(t\) because \(|r|>1\).
In our model, this exponential growth arises not merely from a failure to learn the update in some directions, but from learning an incorrect update that actively amplifies errors.

However, unit or unstable modes do not cause divergence unless they are excited by the context input. When \(\alpha<1\), the context input is confined to a lower-dimensional subspace and may have zero component along these modes (\(c=0\)). In that case, neither error accumulation nor amplification occurs, giving rise to the bounded regime.

These two mechanisms are also consistent with the empirical observations of \citet{ainslie2026eigenanalysis}. 
In autoregressive dynamical-system prediction, they found that models with eigenvalues outside the unit circle (\(|r|>1\)) rapidly amplified errors and produced unphysical trajectories, whereas models with eigenvalues near the unit circle (\(|r|\approx1\)) gradually accumulated prediction errors over time.
Starting from the reasoning dynamics of a linear attention model, our study explains how the system undergoes sharp transitions among the three phases of error accumulation, error amplification, and bounded error depending on the balance between the training CoT budget and context length.

\subsection{Non-monotonic Effects of the Training CoT Budget}
\label{sec:nonmonotonic-cot-budget}

Figure~\ref{fig:training-cot-budget-phases}C shows that, for \(\alpha<1\), increasing the training CoT budget leads to successive transitions from quadratic divergence to exponential divergence, bounded error, and finally GD recovery. 
Notably, at intermediate budgets, adding teacher CoT can worsen long-time error growth.

This non-monotonicity can be understood as follows.
With very little teacher CoT, repeated attention updates add nearly the same context information at each step, causing errors to accumulate.
At intermediate training CoT budgets, the model learns to use its current estimate in the next update, but with short contexts it does not yet learn the correct update rule.
The resulting reasoning dynamics amplify errors from one step to the next, causing exponential growth.
As the training CoT budget increases further, the model learns reliable updates and eventually recovers GD.

A similar non-monotonic pattern has been observed in practical CoT training.
\citet{ren2026rethinking_sft_generalization} report that model performance initially declines and then recovers during supervised fine-tuning with long teacher CoT.
Our theory offers a possible explanation for this behavior by showing that adding teacher CoT can temporarily worsen reasoning performance when short contexts do not sufficiently identify the update rule.

\section{Experiments Beyond the Effective Model}
\label{sec:empirical-extensions}

This section evaluates whether the qualitative behaviors predicted by the effective model persist in more complex architectures. We retain the setup of Section~\ref{sec:model} and vary only the architecture.
Specifically, we evaluate two end-to-end trained setups: a single-head linear-attention model (optimizing \(\mat W_Q,\mat W_K,\mat W_V\)) and a two-head softmax-attention model (optimizing \(\mat W_Q,\mat W_K,\mat W_V,\mat W_O\)).
Both architectures incorporate a context-only mask: the current state serves as the query, whereas the labeled context examples supply the keys and values.
Masking out the state token from the key--value sequence prevents state-only updates and ensures that every inference step depends on the context.
Complete formulations, parameterizations, and optimization details are provided in Appendix~\ref{app:attention-nonmonotonic}.

Figure~\ref{fig:attention-nonmonotonic} reports the test-time generalization error dynamics for linear and softmax attention.
The first observation is that, given sufficient training CoT, both linear and softmax attention maintain stable inference beyond the CoT length used in training.
This suggests that end-to-end trained attention models can learn from finite-length CoT to continue reasoning effectively beyond the training horizon, as discussed in Section~\ref{sec:recovering-gd}.

More importantly, both models exhibit non-monotonic long-horizon performance as the training CoT budget increases: the error remains relatively low at small budgets, grows rapidly with inference depth at intermediate budgets, and then abruptly transitions to stable inference at larger budgets. 
This pattern suggests that the non-monotonicity and phase-transition-like behavior predicted in Section~\ref{sec:nonmonotonic-cot-budget} persist in more complex attention models.

\begin{figure}[t]
  \centering
  \includegraphics[width=0.75\linewidth]{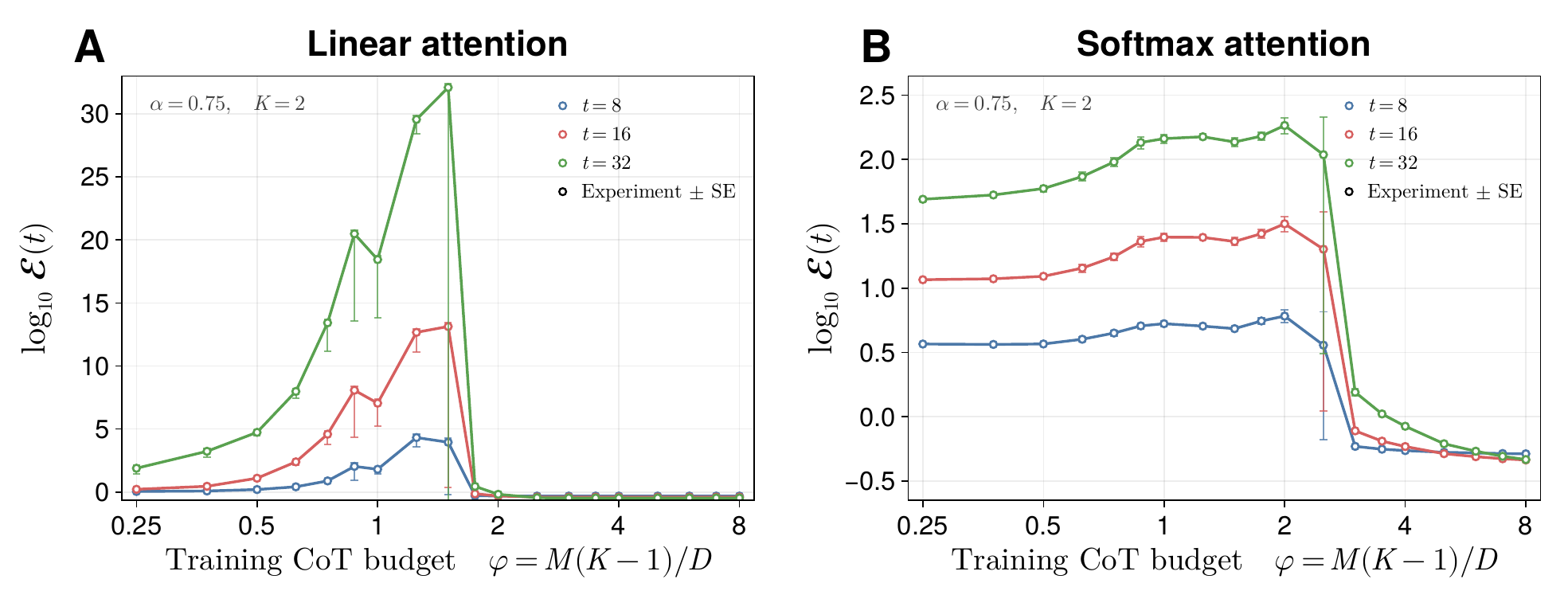}
  \caption{
    \textbf{Training CoT budget produces non-monotonic long-horizon performance in fully trained attention.}
    Numerical experiments of the test-time generalization error for (A) Single-head linear attention with learned Q/K/V and
    (B) Two-head softmax attention with learned Q/K/V/O.
    Parameters: (A,B) \(D=32\), \(\alpha=0.75\), \(K=2\), \(\eta=0.1\), and \(\sigma=0.1\).
    Error bars represent one standard error of the 8 independent experiments per point.
  }
  \label{fig:attention-nonmonotonic}
\end{figure}

Beyond the overall non-monotonicity described above, Figure~\ref{fig:attention-nonmonotonic}A shows a relatively sharp change within the failure region of linear attention. This is consistent with our finding in Section~\ref{sec:two-mechanisms-overthinking} that overthinking takes two distinct forms: error accumulation and error amplification.
For softmax attention, the corresponding change in Figure~\ref{fig:attention-nonmonotonic}B is more gradual, so the failure region is not clearly divided into these two regimes.
To understand this, we write the state update for a fixed context \(Q\) as
\begin{equation}
  \hat{\vb w}_{t+1}
  =F_Q(\hat{\vb w}_t)
  :=\hat{\vb w}_t+\Delta_\theta(Q,\hat{\vb w}_t).
\end{equation}
If the current state contains a small perturbation \(\vb\delta_t\), its one-step evolution can be linearized as
\begin{align}
  \vb\delta_{t+1}
  &:={}
  F_Q(\hat{\vb w}_t+\vb\delta_t)-F_Q(\hat{\vb w}_t)
  \simeq \mat J_t\vb\delta_t,
  &
  \mat J_t
  &:={}
  \left.\frac{\partial F_Q(\vb w)}{\partial\vb w}\right|_{\vb w=\hat{\vb w}_t}.
\end{align}
Here, \(\mat J_t\) describes the one-step evolution of a small state perturbation. 
%For linear attention, \(\mat J_t\) is constant throughout inference for a fixed context.
For linear attention, the Jacobian is constant throughout inference for a fixed context. For softmax attention, it depends on the current state and changes along the inference trajectory. More generally, in nonlinear architectures, error propagation can vary along a single inference trajectory.

\section{Conclusion}
\label{sec:conclusion}

We studied how finite teacher CoT affects test-time reasoning and showed that finite trajectories can teach a GD update beyond the training horizon.
When GD is not fully recovered, the model exhibits bounded error or two distinct forms of overthinking: polynomial error accumulation and exponential error amplification.
In particular, with short contexts, increasing the training CoT budget can temporarily worsen long-horizon performance before ultimately recovering GD.
Experiments with fully trained linear and softmax attention further reproduce these qualitative behaviors beyond the effective model.
These results offer practical guidance for designing training CoT, while extending the theory to more complex models and tasks remains an important direction for future work.

\subsection*{Reproducibility statement}
Detailed derivations are provided in the appendices.
\ifshowreproductioncode
Code for reproducing the numerical experiments is available at \url{https://github.com/taka255/theory_of_cot_in_training}.
\else
Code for reproducing the numerical experiments is provided in the supplementary materials.
\fi

\subsection*{AI use statement}
Generative AI tools assisted with manuscript drafting, editing, translation, paper organization, numerical code, and the exploration and checking of mathematical derivations.
All material was reviewed and verified by the
authors, who take responsibility for the final manuscript.

\ifarxivversion
\subsection*{Acknowledgments and Disclosure of Funding}
\arxivacknowledgements
\fi

\bibliography{ref}
\bibliographystyle{iclr2027_conference}

\clearpage
\appendix
\begin{center}
  {\Large\bfseries Appendix}
\end{center}
\startcontents[appendices]
\section*{Contents}
\printcontents[appendices]{}{1}{\setcounter{tocdepth}{2}}
\section{Derivation of the Effective Parameterization}
\label{app:effective-parameterization}

In this section, we explain the rationale behind the parameter simplification of the effective model in Section~\ref{sec:effective-model}.  

\paragraph{Readout-relevant blocks.}
Write
\begin{equation}
  \mat V:=\mat W_V,
  \qquad
  \mat W:=\mat W_K^\top\mat W_Q,
\end{equation}
and partition both matrices into blocks of dimensions \((D,1,D,1)\):
\begin{equation}
  \mat V=
  \begin{pmatrix}
    \mat V_{11}&\mat V_{12}&\mat V_{13}&\mat V_{14}\\
    \mat V_{21}&V_{22}&\mat V_{23}&V_{24}\\
    \mat V_{31}&\mat V_{32}&\mat V_{33}&\mat V_{34}\\
    \mat V_{41}&V_{42}&\mat V_{43}&V_{44}
  \end{pmatrix},
  \qquad
  \mat W=
  \begin{pmatrix}
    \mat W_{11}&\mat W_{12}&\mat W_{13}&\mat W_{14}\\
    \mat W_{21}&W_{22}&\mat W_{23}&W_{24}\\
    \mat W_{31}&\mat W_{32}&\mat W_{33}&\mat W_{34}\\
    \mat W_{41}&W_{42}&\mat W_{43}&W_{44}
  \end{pmatrix}.
  \label{eq:appendix-unrestricted-attention-blocks}
\end{equation}
Here every block has the conformable dimension determined by its row and
column; in particular, \(\mat V_{32},\mat V_{34},\mat W_{14},\mat W_{34}\)
are column vectors, and \(\mat W_{23},\mat W_{43}\) are row vectors.

For notational simplicity, suppress the task and step indices and write a
context token and the state token as
\begin{equation}
  \vb c_l:=
  \begin{pmatrix}\vb x_l\\y_l\\\vb0_D\\0\end{pmatrix},
  \qquad
  \vb z(\vb s):=
  \begin{pmatrix}\vb0_D\\0\\\vb s\\1\end{pmatrix},
  \qquad
  \mat E=(\vb c_1,\ldots,\vb c_L,\vb z(\vb s)).
\end{equation}
The state readout selects only the third block row of \(\mat V\).  
Moreover, the query \(\vb z(\vb s)\) has support only in its third and fourth blocks, so only the third and fourth block columns of \(\mat W\) enter the attention scores.  
Direct multiplication gives
\begin{align}
  \mathsf{Readout}\ab(\mathsf{Atten}(\mat E;\theta)\ab)
  ={}&\vb s+\frac1L
  \ab(\mat V_{31}\mat X+\mat V_{32}\vb y^\top\ab)
  \Big[
    \mat X^\top\ab(\mat W_{13}\vb s+\mat W_{14}\ab)
    +\vb y\ab(\mat W_{23}\vb s+W_{24}\ab)
  \Big]
  \nonumber\\
  &+\frac1L
  \ab(\mat V_{33}\vb s+\mat V_{34}\ab)
  \Big[
    \vb s^\top\mat W_{33}\vb s
    +\vb s^\top\mat W_{34}
    +\mat W_{43}\vb s+W_{44}
  \Big].
  \label{eq:appendix-unrestricted-attention-output}
\end{align}
Consequently, all blocks outside the third block row of \(\mat V\), and the
first two block columns of \(\mat W\), are exactly irrelevant to the state
readout.  They may be arbitrary, which gives the \(\ast\) blocks in
Equation~\eqref{eq:effective-attention-parameterization}.

\paragraph{Consequences of rotational symmetry.}
The data distribution is rotationally symmetric in the feature space.
Motivated by this symmetry, we assume that the fixed blocks do not introduce preferred directions.  
The blocks
\(\mat V_{32},\mat V_{34},\mat W_{14},\mat W_{34}\) would otherwise
introduce fixed vectors in \(\R^D\), while \(\mat W_{23},\mat W_{43}\)
would select fixed directions from the state.  We therefore set
\begin{equation}
  \mat V_{32}=\mat V_{34}=\mat W_{14}=\mat W_{34}=\vb0_D,
  \qquad
  \mat W_{23}=\mat W_{43}=\vb0_D^\top.
  \label{eq:appendix-symmetry-zero-blocks}
\end{equation}
For the same reason, we take the fixed square blocks to be proportional to \(\mat I_D\), while leaving \(\mat W_{13}=\mat A\) unrestricted.  Thus,
\begin{equation}
  \mat V_{31}=a\mat I_D,
  \qquad
  \mat V_{33}=b\mat I_D,
  \qquad
  \mat W_{33}=d\mat I_D,
  \qquad
  W_{24}=c,
  \qquad
  W_{44}=e.
  \label{eq:appendix-symmetry-isotropic-blocks}
\end{equation}

Substitution into Equation~\eqref{eq:appendix-unrestricted-attention-output}
gives
\begin{equation}
  \mathsf{Readout}\ab(\mathsf{Atten}(\mat E;\theta)\ab)
  =\vb s
  +\frac{a}{L}\mat X\ab(\mat X^\top\mat A\vb s+c\vb y\ab)
  +\frac{b}{L}\vb s\ab(d\|\vb s\|_2^2+e\ab).
  \label{eq:appendix-rotation-compatible-update}
\end{equation}

\paragraph{Scale convention and the label coefficient.}
The parameters in Equation~\eqref{eq:appendix-rotation-compatible-update}
appear only through the products \(a\mat A\), \(ac\), \(bd\), and \(be\).
We fix this scalar gauge by absorbing \(a\) into \(\mat A\) and \(c\), and
set \(a=1\).  The value of \(b\) becomes irrelevant after the state-only
terms are removed below; we choose \(b=1\), giving
\(\mat V_{31}=\mat V_{33}=\mat I_D\).

The scalar \(c\) need not be fixed as an additional assumption.  If it is
included in the convex learning problem, the transition from
\(\vb s_{\mu,0}=\vb0_D\) satisfies
\begin{equation}
  \hat{\vb s}_{\mu,1}=\frac{c}{L}\mat X_\mu\vb y_\mu,
  \qquad
  \vb s_{\mu,1}=\frac{\eta}{L}\mat X_\mu\vb y_\mu.
  \label{eq:appendix-initial-transition-identifies-c}
\end{equation}
The teacher parameters interpolate all training transitions, so the
\(\lambda\to0\) solution has zero training error.  Equation~\eqref{eq:appendix-initial-transition-identifies-c}
then requires \((c-\eta)\mat X_\mu\vb y_\mu=\vb0_D\).  
Since \(\mat X_\mu\vb y_\mu\ne\vb0_D\) almost surely, every zero-loss solution has
\begin{equation}
  c=\eta.
  \label{eq:appendix-label-coefficient}
\end{equation}
Thus fixing this scalar from the outset entails no loss of generality for the \(\lambda\to0\) results.

\paragraph{Excluding context-free inference.}
The last term of Equation~\eqref{eq:appendix-rotation-compatible-update}
depends only on \(\vb s\) and is independent of the context
\((\mat X,\vb y)\).  We therefore set
\begin{equation}
  d=e=0
  \label{eq:appendix-context-grounding}
\end{equation}
to exclude context-independent state updates.  Combining
Equations~\eqref{eq:appendix-symmetry-zero-blocks},
\eqref{eq:appendix-label-coefficient}, and
\eqref{eq:appendix-context-grounding} yields
\begin{equation}
  \mat V=
  \begin{pmatrix}
    \ast&\ast&\ast&\ast\\
    \ast&\ast&\ast&\ast\\
    \mat I_D&\vb0_D&\mat I_D&\vb0_D\\
    \ast&\ast&\ast&\ast
  \end{pmatrix},
  \qquad
  \mat W=
  \begin{pmatrix}
    \ast&\ast&\mat A&\vb0_D\\
    \ast&\ast&\vb0_D^\top&\eta\\
    \ast&\ast&\mat0_{D\times D}&\vb0_D\\
    \ast&\ast&\vb0_D^\top&0
  \end{pmatrix}.
\end{equation}
The unrestricted attention output therefore reduces to
\begin{equation}
  \mathsf{Readout}\ab(\mathsf{Atten}(\mat E;\theta)\ab)
  =\vb s+\frac1L\mat X
  \ab(\mat X^\top\mat A\vb s+\eta\vb y\ab),
\end{equation}
which is Equation~\eqref{eq:effective-update}.  In particular,
\(\mat A=-\eta\mat I_D\) recovers the teacher GD update exactly.

\section{Summary of the Parameter Regimes}
\label{app:phase-summary}

Table~\ref{tab:parameter-regime-summary} summarizes the finite-time and
long-time behavior in every open region of the \((\alpha,\varphi)\) plane.

\begin{table}[h]
  \centering
  \caption{Summary of the parameter regimes.}
  \label{tab:parameter-regime-summary}
  \small
  \begin{tabular}{@{}p{0.20\linewidth}p{0.19\linewidth}p{0.20\linewidth}p{0.27\linewidth}@{}}
    \toprule
    Regime
      & Finite-time behavior (Result~\ref{res:finite-depth-rich-context})
      & Long-time behavior (Result~\ref{res:long-time-phases})
      & Derivation\\
    \midrule
    \(\alpha>1,\ \varphi>1\)
      & GD
      & GD
      & Appendix~\ref{app:proof-phase-one} \\
    \(\alpha>1,\ 0<\varphi<1\)
      & Projected GD
      & \(\Theta(t^2)\)
      & Appendix~\ref{app:proof-phase-two} \\
    \(0<\alpha<1,\ 0<\varphi<\varphi_-\)
      & --
      & \(\Theta(t^2)\)
      & Proposition~\ref{prop:long-time-cavity-phases} \\
    \(0<\alpha<1,\ \varphi_-<\varphi<\varphi_+\)
      & --
      & \(\exp(\Theta(\eta t))\)
      & Proposition~\ref{prop:long-time-cavity-phases} \\
    \(0<\alpha<1,\ \varphi_+<\varphi<1/\alpha\)
      & --
      & \(O(1)\)
      & Proposition~\ref{prop:long-time-cavity-phases} \\
    \(0<\alpha<1,\ \alpha\varphi>1\)
      & GD
      & GD
      & Proposition~\ref{prop:rank-deficient-gd-recovery} \\
    \bottomrule
  \end{tabular}
\end{table}

The two boundaries for \(0<\alpha<1\) are derived in
Propositions~\ref{prop:lower-boundary} and \ref{prop:upper-boundary}.

\section{The \(\alpha>1,\ \varphi>1\) Case}
\label{app:proof-phase-one}

In this section, we derive the finite-depth generalization error in the GD-recovery regime of Result~\ref{res:finite-depth-rich-context}, especially in the $\alpha>1$ case.
The projected-GD error is evaluated separately in Appendix~\ref{app:proof-phase-two}.
We write the task and inference covariances as
\begin{equation}
  \mat H_\mu:=\frac{\mat X_\mu\mat X_\mu^\top}{L},
  \qquad
  \mat H^{\mathrm{in}}:=\frac{\mat X^{\mathrm{in}}(\mat X^{\mathrm{in}})^\top}{L}.
  \label{eq:appendix-task-covariances}
\end{equation}

First, we verify that the GD parameters satisfy \(\mathcal L_{\mathrm{eff}}(\mat A_\star)=0\).
This is a direct substitution into the model update, after which only uniqueness remains to be shown.

\begin{proposition}[The GD parameters satisfy \(\mathcal L_{\mathrm{eff}}=0\)]
\label{prop:gd-zero-error}
Suppose \(\lambda\to0\).
For every \(\alpha>0\), the parameter \(\mat A_\star=-\eta\mat I_D\) satisfies
\begin{equation}
  \mathcal L_{\mathrm{eff}}(\mat A_\star)
  =\frac{1}{MK}
  \sum_{\mu=1}^M\sum_{k=0}^{K-1}
  \ab\|\hat{\vb{s}}_{\mu, k+1}
    -\vb s_{\mu,k+1}\ab\|_2^2
  \bigg|_{\mat A=\mat A_\star}
  ={}
  0.
  \label{eq:gd-parameters-zero-error}
\end{equation}
\end{proposition}

\begin{proof}
Substituting \(\mat A_\star=-\eta\mat I_D\) into the model update in Equation~\eqref{eq:effective-update} gives
\begin{align}
  \hat{\vb s}_{\mu,k+1}
  &=\vb s_{\mu,k}
    +\frac1L\mat X_\mu
      \left[-\eta\mat X_\mu^\top\vb s_{\mu,k}+\eta\vb y_\mu\right]\\
  &=\vb s_{\mu,k}
    -\frac{\eta}{L}\mat X_\mu
      \left(\mat X_\mu^\top\vb s_{\mu,k}-\vb y_\mu\right)\\
  &=\vb s_{\mu,k+1},
  \label{eq:gd-parameter-substitution}
\end{align}
where the last equality is the teacher update in Equation~\eqref{eq:teacher-gd-trajectory}.
Substituting Equation~\eqref{eq:gd-parameter-substitution} into Equation~\eqref{eq:effective-regularized-loss} proves Equation~\eqref{eq:gd-parameters-zero-error} in the limit \(\lambda\to0\).
\end{proof}

Proposition~\ref{prop:gd-zero-error} establishes that \(\mathcal L_{\mathrm{eff}}(\mat A_\star)=0\).
We now assume \(\alpha>1\) and show that \(\mat A_\star\) is the unique matrix satisfying \(\mathcal L_{\mathrm{eff}}(\mat A)=0\) when \(\varphi>1\).
Because every task covariance is then invertible, the proof reduces to the rank of one matrix containing all training states.

\begin{proposition}[Uniqueness for full-rank contexts]
\label{prop:gd-uniqueness}
Suppose \(\alpha>1\), \(\varphi>1\), and \(\lambda\to0\).
Then \(\mat A_\star\) is the unique solution that satisfies $\mathcal{L}_{\text{eff}}(\mat{A}) = 0$, and thus
\begin{equation}
  \hat{\mat A}=-\eta\mat I_D.
  \label{eq:full-rank-gd-recovery-condition}
\end{equation}
\end{proposition}

\begin{proof}
Subtracting the teacher update in Equation~\eqref{eq:teacher-gd-trajectory} from the model update in Equation~\eqref{eq:effective-update} gives
\begin{equation}
  \hat{\vb s}_{\mu,k+1}-\vb s_{\mu,k+1}
  =\mat H_\mu(\mat A+\eta\mat I_D)\vb s_{\mu,k}.
  \label{eq:proof-training-residual}
\end{equation}
Since \(\vb s_{\mu,0}=\vb0\), the \(k=0\) term in \(\mathcal L_{\mathrm{eff}}(\mat A)\) vanishes for every \(\mat A\).
Equation~\eqref{eq:proof-training-residual} therefore shows that \(\mathcal L_{\mathrm{eff}}(\mat A)=0\) requires
\begin{equation}
  \mat H_\mu(\mat A+\eta\mat I_D)\vb s_{\mu,k}=\vb0_D
  \qquad(\mu=1,\ldots,M,\ k=1,\ldots,K-1).
  \label{eq:proof-interpolation-constraints}
\end{equation}
Because \(\alpha>1\), every \(\mat H_\mu\) is invertible almost surely.
Collect all noninitial teacher states in
\begin{equation}
  \mat S:=
  (\vb s_{1,1},\ldots,\vb s_{1,K-1},\ldots,
   \vb s_{M,1},\ldots,\vb s_{M,K-1}).
  \label{eq:full-rank-state-matrix}
\end{equation}
Then Equation~\eqref{eq:proof-interpolation-constraints} is equivalent to
\begin{equation}
  \mathcal L_{\mathrm{eff}}(\mat A)=0
  \quad\Longleftrightarrow\quad
  (\mat A+\eta\mat I_D)\mat S=\mat0.
  \label{eq:full-rank-identification-constraints}
\end{equation}
The teacher recursion in Equation~\eqref{eq:teacher-gd-trajectory} yields
\begin{equation}
  \vb s_{\mu,k}
  =\frac1L q_k(\mat H_\mu)\mat X_\mu\vb y_\mu,
  \label{eq:proof-krylov-states}
\end{equation}
where $q_k(x)$ is defined in Equation~\eqref{eq:mp-average-gd-filter}.
Write
\begin{equation}
  \mat H_\mu
  =\mat U_\mu\operatorname{diag}(x_{\mu,1},\ldots,x_{\mu,D})\mat U_\mu^\top.
  \label{eq:proof-task-eigendecomposition}
\end{equation}
Equation~\eqref{eq:proof-krylov-states} then gives
\begin{equation}
  \mat U_\mu^\top
  (\vb s_{\mu,1},\ldots,\vb s_{\mu,K-1})
  =\operatorname{diag}\!\left(\frac1L\mat U_\mu^\top\mat X_\mu\vb y_\mu\right)
  \begin{pmatrix}
    q_1(x_{\mu,1})&\cdots&q_{K-1}(x_{\mu,1})\\
    \vdots&&\vdots\\
    q_1(x_{\mu,D})&\cdots&q_{K-1}(x_{\mu,D})
  \end{pmatrix}.
  \label{eq:proof-polynomial-evaluation-matrix}
\end{equation}
The polynomial \(q_k\) has degree \(k-1\) and leading coefficient \((-1)^{k-1}\eta^k\).
Therefore, the determinant of the first \(K-1\) rows of the polynomial matrix in Equation~\eqref{eq:proof-polynomial-evaluation-matrix} is
\begin{equation}
  \det\!\left[q_k(x_{\mu,i})\right]_{i,k=1}^{K-1}
  =\left[\prod_{k=1}^{K-1}(-1)^{k-1}\eta^k\right]
   \prod_{1\le i<j\le K-1}(x_{\mu,j}-x_{\mu,i}).
  \label{eq:proof-vandermonde-determinant}
\end{equation}
The positive Gaussian-Wishart eigenvalues are distinct almost surely, and every entry of \(\mat U_\mu^\top\mat X_\mu\vb y_\mu/L\) is nonzero almost surely.
Hence Equation~\eqref{eq:proof-vandermonde-determinant} is nonzero, and each task contributes exactly \(K-1\) linearly independent states.
Because the tasks are independent and Gaussian, their state spans are independent random subspaces and are in general position almost surely.
Consequently,
\begin{equation}
  \operatorname{rank}\mat S=\min\{D,M(K-1)\},
  \qquad
  \frac{\operatorname{rank}\mat S}{D}\longrightarrow\min\{1,\varphi\}.
  \label{eq:training-subspace-dimension}
\end{equation}
When \(\varphi>1\), Equation~\eqref{eq:training-subspace-dimension} gives \(\operatorname{rank}\mat S=D\).
Equation~\eqref{eq:full-rank-identification-constraints} then gives \(\mat A=-\eta\mat I_D=\mat A_\star\).
Hence \(\mat A_\star\) is the unique matrix satisfying \(\mathcal L_{\mathrm{eff}}(\mat A)=0\), and the \(\lambda\to0\) estimator obeys \(\hat{\mat A}=-\eta\mat I_D\).
\end{proof}

Propositions~\ref{prop:gd-zero-error} and \ref{prop:gd-uniqueness} establish GD recovery for \(\alpha>1\) and \(\varphi>1\).
Finally, we compute the generalization error after GD recovery.
This calculation applies whenever \(\hat{\mat A}=-\eta\mat I_D\).

\begin{proposition}[Generalization error after GD recovery]
\label{prop:gd-recovery-risk}
Suppose \(\lambda\to0\) and \(\hat{\mat A}=-\eta\mat I_D\).
For every fixed test-time depth \(t\),
\begin{equation}
  \mathcal E(t)=\mathcal E_{\mathrm{GD}}(t)
  =\left\langle(1-xq_t)^2\right\rangle_\alpha
   +\frac{\sigma^2}{\alpha}\left\langle xq_t^2\right\rangle_\alpha.
  \label{eq:proof-recovered-gd-risk}
\end{equation}
\end{proposition}

\begin{proof}
Define
\begin{equation}
  \mat Q_t:=\eta\sum_{j=0}^{t-1}
      (\mat I_D+\mat H^{\mathrm{in}}\hat{\mat A})^j.
  \label{eq:proof-general-response}
\end{equation}
Iterating the learned update in Equation~\eqref{eq:learned-iteration} and substituting the test labels gives
\begin{equation}
  \hat{\vb w}_t-\vb w^{\mathrm{in}}
  =(\mat Q_t\mat H^{\mathrm{in}}-\mat I_D)\vb w^{\mathrm{in}}
   +\frac1L\mat Q_t\mat X^{\mathrm{in}}\vb\varepsilon^{\mathrm{in}}.
  \label{eq:proof-response-and-error}
\end{equation}
The two terms in Equation~\eqref{eq:proof-response-and-error} are conditionally independent and centered.
Their covariances give
\begin{align}
  \E\!\left[\|\hat{\vb w}_t-\vb w^{\mathrm{in}}\|_2^2
       \,\middle|\,\hat{\mat A},\mat X^{\mathrm{in}}\right]
  &=1-\frac2D\operatorname{tr}(\mat Q_t\mat H^{\mathrm{in}})
    +\frac1D\|\mat Q_t\mat H^{\mathrm{in}}\|_{\mathrm F}^2
  +\frac{\sigma^2}{L}
    \operatorname{tr}(\mat Q_t\mat H^{\mathrm{in}}\mat Q_t^\top).
  \label{eq:proof-conditional-risk}
\end{align}
Substituting \(\hat{\mat A}=-\eta\mat I_D\) into Equation~\eqref{eq:proof-general-response} gives
\begin{equation}
  \mat Q_t
  =\eta\sum_{j=0}^{t-1}(\mat I_D-\eta\mat H^{\mathrm{in}})^j
  =q_t(\mat H^{\mathrm{in}}).
  \label{eq:proof-full-gd-response}
\end{equation}
Let \(x_{1,D},\ldots,x_{D,D}\) be the eigenvalues of
\(\mat H^{\mathrm{in}}\).
For fixed \(t\), \(q_t\) is a polynomial, so the Marchenko--Pastur law
applies term by term in Equation~\eqref{eq:proof-conditional-risk}.
Using Equation~\eqref{eq:proof-full-gd-response},
\begin{align}
  \frac1D\operatorname{tr}(\mat Q_t\mat H^{\mathrm{in}})
  &=\frac1D\sum_{i=1}^D x_{i,D}q_t(x_{i,D})
  \longrightarrow \left\langle xq_t\right\rangle_\alpha,
  \label{eq:proof-gd-linear-trace}\\
  \frac1D\|\mat Q_t\mat H^{\mathrm{in}}\|_{\mathrm F}^2
  &=\frac1D\operatorname{tr}\!\left[
      (\mat H^{\mathrm{in}})^2q_t(\mat H^{\mathrm{in}})^2\right]\\
  &=\frac1D\sum_{i=1}^D x_{i,D}^2q_t(x_{i,D})^2
  \longrightarrow \left\langle x^2q_t^2\right\rangle_\alpha,
  \label{eq:proof-gd-quadratic-trace}\\
  \frac{\sigma^2}{L}
    \operatorname{tr}(\mat Q_t\mat H^{\mathrm{in}}\mat Q_t^\top)
  &=\sigma^2\frac DL\frac1D
    \operatorname{tr}\!\left[
      \mat H^{\mathrm{in}}q_t(\mat H^{\mathrm{in}})^2\right]\\
  &=\sigma^2\frac DL\frac1D
      \sum_{i=1}^D x_{i,D}q_t(x_{i,D})^2
  \longrightarrow
    \frac{\sigma^2}{\alpha}\left\langle xq_t^2\right\rangle_\alpha.
  \label{eq:proof-gd-noise-trace}
\end{align}
Substitution into Equation~\eqref{eq:proof-conditional-risk} gives
\begin{align}
  \mathcal E_{\mathrm{GD}}(t)
  &=1-2\left\langle xq_t\right\rangle_\alpha
    +\left\langle x^2q_t^2\right\rangle_\alpha
    +\frac{\sigma^2}{\alpha}\left\langle xq_t^2\right\rangle_\alpha\\
  &=\left\langle(1-xq_t)^2\right\rangle_\alpha
    +\frac{\sigma^2}{\alpha}\left\langle xq_t^2\right\rangle_\alpha,
\end{align}
which proves Equation~\eqref{eq:proof-recovered-gd-risk}.
\end{proof}

This completes the $\alpha>1$ and $\varphi>1$ case.

\section{The \(\alpha>1,\ 0<\varphi<1\) Case}
\label{app:proof-phase-two}

In this section, we compute the generalization error for \(\alpha>1\) and \(0<\varphi<1\).
First, we state the finite-time error and derive both the learned matrix and its risk.

\begin{proposition}[Finite-time error of projected GD]
\label{prop:projected-gd-risk}
Suppose \(\alpha>1\), \(0<\varphi<1\), and \(\lambda\to0\).
For every fixed test-time depth \(t\), the quantities defined in Equation~\eqref{eq:mp-average-gd-filter} give
\begin{align}
  \mathcal E(t)
  ={}&1-\varphi+\varphi\langle(1-xq_t)^2\rangle_{\alpha/\varphi}
  +\frac{\varphi[2(1-\varphi)+\sigma^2]}{\alpha}
       \langle xq_t^2\rangle_{\alpha/\varphi}\notag\\
  &-\frac{2(1-\varphi)}{\alpha}\kappa_{1,t}
  +\frac{(1-\varphi)(1-\varphi+\sigma^2)}{\alpha^2}\kappa_{2,t}
  +\frac{1-\varphi}{\alpha^2}\kappa_{1,t}^2.
  \label{eq:proof-projected-gd-risk}
\end{align}
\end{proposition}

\begin{proof}
Define the projector onto the training-state span:
\begin{equation}
  \mat P:=\mat S\mat S^\dagger.
  \label{eq:proof-projected-state-projector}
\end{equation}
Every noninitial training state is a column of \(\mat S\), so
\(\mat P\vb s_{\mu,k}=\vb s_{\mu,k}\) for \(k=1,\ldots,K-1\).
Using Equation~\eqref{eq:proof-training-residual},
\begin{equation}
  \left.
  (\hat{\vb s}_{\mu,k+1}-\vb s_{\mu,k+1})
  \right|_{\mat A=-\eta\mat P}
  =\eta\mat H_\mu(\mat I_D-\mat P)\vb s_{\mu,k}
  =\vb0_D.
  \label{eq:proof-projected-zero-residual}
\end{equation}
The \(k=0\) residual also vanishes because \(\vb s_{\mu,0}=\vb0_D\).
Consequently,
\begin{equation}
  \mathcal L_{\mathrm{eff}}(-\eta\mat P)
  =\frac{\lambda\eta^2}{D}\|\mat P\|_{\mathrm F}^2
  =\frac{\lambda\eta^2}{D}\operatorname{tr}\mat P
  \longrightarrow0
  \qquad(\lambda\to0).
  \label{eq:proof-projected-zero-loss}
\end{equation}
Multiplying Equation~\eqref{eq:full-rank-identification-constraints} by \(\mat S^\dagger\) gives
\begin{equation}
  \mat A\mat P=-\eta\mat P.
  \label{eq:proof-projected-parameter-condition}
\end{equation}
Every matrix satisfying Equation~\eqref{eq:proof-projected-parameter-condition} obeys
\begin{align}
  \mat A
  &=-\eta\mat P+\mat A(\mat I_D-\mat P),\\
  \|\mat A\|_{\mathrm F}^2
  &=\eta^2\operatorname{tr}\mat P
    +\|\mat A(\mat I_D-\mat P)\|_{\mathrm F}^2.
  \label{eq:proof-projected-norm-decomposition}
\end{align}
Because \(\lambda\to0\) selects the minimum-Frobenius-norm matrix satisfying \(\mathcal L_{\mathrm{eff}}(\mat A)=0\), Equation~\eqref{eq:proof-projected-norm-decomposition} gives
\begin{equation}
  \hat{\mat A}=-\eta\mat P.
  \label{eq:proof-projected-update}
\end{equation}
Let \(r:=M(K-1)\).
Because \(0<\varphi<1\), Equation~\eqref{eq:training-subspace-dimension} gives \(\operatorname{rank}\mat P=r\) and \(r/D\to\varphi\).
The inference data are independent of \(\mat P\), so rotational invariance allows us to choose a basis in which
\begin{equation}
  \mat P=
  \begin{pmatrix}\mat I_r&\mat0\\\mat0&\mat0\end{pmatrix},
  \qquad
  \mat X^{\mathrm{in}}=
  \begin{pmatrix}\mat X_1\\\mat X_2\end{pmatrix},
  \qquad
  \mat B:=\frac{\mat X_1\mat X_1^\top}{L},
  \label{eq:proof-projected-blocks}
\end{equation}
where \(\mat X_1\in\R^{r\times L}\) and \(\mat X_2\in\R^{(D-r)\times L}\) are independent standard Gaussian matrices.
Substituting \(\hat{\mat A}=-\eta\mat P\) into Equation~\eqref{eq:proof-general-response} gives
\begin{equation}
  \mat Q_t=q_t(\mat H^{\mathrm{in}}\mat P).
  \label{eq:proof-projected-response}
\end{equation}
The identity
\begin{equation}
  (\mat H^{\mathrm{in}}\mat P)^j\mat X^{\mathrm{in}}
  =\mat X^{\mathrm{in}}
   \left[\frac{(\mat X^{\mathrm{in}})^\top\mat P\mat X^{\mathrm{in}}}{L}\right]^j
  \label{eq:proof-projected-power-identity}
\end{equation}
and the block decomposition in Equation~\eqref{eq:proof-projected-blocks} give
\begin{align}
  \mat Q_t\mat X^{\mathrm{in}}
  =\mat X^{\mathrm{in}}
       q_t\!\left(\frac{\mat X_1^\top\mat X_1}{L}\right)
  =\begin{pmatrix}q_t(\mat B)\mat X_1\\\mat X_2\mat K_t\end{pmatrix},
  \label{eq:proof-projected-source-response}
\end{align}
where we introduced
\begin{align}
  \mat K_t:=q_t\!\left(\frac{\mat X_1^\top\mat X_1}{L}\right).
\end{align}
Multiplying Equation~\eqref{eq:proof-projected-source-response} by \((\mat X^{\mathrm{in}})^\top/L\) gives
\begin{equation}
  \mat Q_t\mat H^{\mathrm{in}}
  =\begin{pmatrix}
      \mat Bq_t(\mat B)&q_t(\mat B)\mat X_1\mat X_2^\top/L\\
      \mat X_2\mat X_1^\top q_t(\mat B)/L
        &\mat X_2\mat K_t\mat X_2^\top/L
    \end{pmatrix}.
  \label{eq:proof-projected-signal-response}
\end{equation}

We next average the \(\mat X_2\) blocks in Equations~\eqref{eq:proof-projected-source-response} and \eqref{eq:proof-projected-signal-response} conditional on \(\mat X_1\).
The matrix \(\mat K_t\) is symmetric and independent of \(\mat X_2\).
For independent standard Gaussian rows \(\vb z_i\) and \(\vb z_j\) of \(\mat X_2\), Wick's identity gives
\begin{align}
  \E(\vb z_i^\top\mat K_t\vb z_i)
  &=\operatorname{tr}\mat K_t,\\
  \E(\vb z_i^\top\mat K_t\vb z_i)^2
  &=2\operatorname{tr}(\mat K_t^2)+(\operatorname{tr}\mat K_t)^2,\\
  \E(\vb z_i^\top\mat K_t\vb z_j)^2
  &=\operatorname{tr}(\mat K_t^2)\qquad(i\ne j).
  \label{eq:proof-wick-row-moments}
\end{align}
Summing the diagonal and off-diagonal row pairs in Equation~\eqref{eq:proof-wick-row-moments} gives
\begin{align}
  \E_{\mat X_2}\operatorname{tr}(\mat X_2\mat K_t\mat X_2^\top)
  &=(D-r)\operatorname{tr}\mat K_t,\\
  \E_{\mat X_2}\|\mat X_2\mat K_t\mat X_2^\top\|_{\mathrm F}^2
  &=(D-r)(D-r+1)\operatorname{tr}(\mat K_t^2)
     +(D-r)(\operatorname{tr}\mat K_t)^2.
  \label{eq:proof-wick-fourth-moment}
\end{align}
The off-diagonal block in Equation~\eqref{eq:proof-projected-signal-response} satisfies
\begin{align}
  \E_{\mat X_2}
     \left\|q_t(\mat B)\frac{\mat X_1\mat X_2^\top}{L}\right\|_{\mathrm F}^2
  =\frac{D-r}{L^2}
       \operatorname{tr}\!\left[q_t(\mat B)\mat X_1\mat X_1^\top q_t(\mat B)\right]
  =\frac{D-r}{L}\operatorname{tr}\!\left[\mat Bq_t(\mat B)^2\right].
  \label{eq:proof-projected-off-diagonal-moment}
\end{align}
The source response in Equation~\eqref{eq:proof-projected-source-response} satisfies
\begin{align}
  \E_{\mat X_2}\|\mat Q_t\mat X^{\mathrm{in}}\|_{\mathrm F}^2
  &=\|q_t(\mat B)\mat X_1\|_{\mathrm F}^2
      +\E_{\mat X_2}\|\mat X_2\mat K_t\|_{\mathrm F}^2\\
  &=L\operatorname{tr}\!\left[\mat Bq_t(\mat B)^2\right]
      +(D-r)\operatorname{tr}(\mat K_t^2).
  \label{eq:proof-projected-noise-moment}
\end{align}
Substituting Equations~\eqref{eq:proof-wick-fourth-moment}, \eqref{eq:proof-projected-off-diagonal-moment}, and \eqref{eq:proof-projected-noise-moment} into Equation~\eqref{eq:proof-conditional-risk} gives
\begin{align}
  \E\!\left[\|\hat{\vb w}_t-\vb w^{\mathrm{in}}\|_2^2
       \,\middle|\,\mat X_1\right]
  ={}&1-\frac2D\left[
       \operatorname{tr}(\mat Bq_t(\mat B))
       +\frac{D-r}{L}\operatorname{tr}\mat K_t\right]\notag\\
    &+\frac1D\operatorname{tr}\!\left[\mat B^2q_t(\mat B)^2\right]
      +\left[\frac{2(D-r)}{DL}
              +\frac{\sigma^2}{L}\right]
         \operatorname{tr}\!\left[\mat Bq_t(\mat B)^2\right]\notag\\
    &+\frac{(D-r)(D-r+1+\sigma^2D)}{DL^2}
         \operatorname{tr}(\mat K_t^2)
      +\frac{D-r}{DL^2}(\operatorname{tr}\mat K_t)^2.
  \label{eq:proof-projected-finite-dimensional-risk}
\end{align}

Finally, we evaluate the traces in Equation~\eqref{eq:proof-projected-finite-dimensional-risk} by the Marchenko--Pastur law.
The block covariance \(\mat B\) has aspect ratio
\begin{equation}
  \frac{L}{r}
  =\frac{L/D}{r/D}
  \longrightarrow\frac{\alpha}{\varphi}.
  \label{eq:proof-projected-aspect-ratio}
\end{equation}
Therefore, for every fixed polynomial \(f\),
\begin{equation}
  \frac1D\operatorname{tr}f(\mat B)
  =\frac rD\,\frac1r\operatorname{tr}f(\mat B)
  \longrightarrow\varphi\langle f\rangle_{\alpha/\varphi}.
  \label{eq:proof-projected-mp-limit}
\end{equation}
The matrices \(\mat X_1^\top\mat X_1/L\) and \(\mat B\) have the same \(r\) nonzero eigenvalues.
The matrix \(\mat X_1^\top\mat X_1/L\) also has \(L-r\) zero eigenvalues, and Equation~\eqref{eq:mp-average-gd-filter} gives \(q_t(0)=\eta t\).
Hence, for \(j=1,2\),
\begin{align}
  \frac1D\operatorname{tr}(\mat K_t^j)
  &=\frac{L-r}{D}(\eta t)^j
       +\frac rD\,\frac1r\operatorname{tr}(q_t(\mat B)^j)\\
  &\longrightarrow(\alpha-\varphi)(\eta t)^j
       +\varphi\langle q_t^j\rangle_{\alpha/\varphi}
   =\kappa_{j,t}.
  \label{eq:proof-projected-companion-traces}
\end{align}
Substituting Equations~\eqref{eq:proof-projected-mp-limit} and \eqref{eq:proof-projected-companion-traces} into Equation~\eqref{eq:proof-projected-finite-dimensional-risk} gives Equation~\eqref{eq:proof-projected-gd-risk}, which is Equation~\eqref{eq:phase-two-gd-risk} in Result~\ref{res:finite-depth-rich-context}.
\end{proof}

Proposition~\ref{prop:projected-gd-risk} gives the exact fixed-time error without a cavity approximation.
We now derive its long-time growth for \(\alpha>1\).

\begin{proposition}[Quadratic long-time growth of projected GD]
\label{prop:projected-gd-long-time}
Suppose \(\alpha>1\), \(0<\varphi<1\), and
\begin{equation}
  0<\eta<\frac{2}{(1+\sqrt{\varphi/\alpha})^2}.
  \label{eq:phase-two-stable-step-size}
\end{equation}
Then
\begin{equation}
  \lim_{t\to\infty}\frac{\mathcal E(t)}{t^2}
  =\frac{\eta^2(1-\varphi)(\alpha-\varphi)
      (\alpha+1-2\varphi+\sigma^2)}{\alpha^2}>0.
  \label{eq:phase-two-gd-risk-growth}
\end{equation}
\end{proposition}

\begin{proof}
Because \(\alpha/\varphi>1\), the Marchenko--Pastur support is separated from zero.
Under Equation~\eqref{eq:phase-two-stable-step-size}, the nonzero modes are stable, and therefore
\begin{equation}
  \langle q_t\rangle_{\alpha/\varphi}=O(1),
  \qquad
  \langle q_t^2\rangle_{\alpha/\varphi}=O(1),
  \qquad t\to\infty.
  \label{eq:proof-projected-bounded-moments}
\end{equation}
Equations~\eqref{eq:mp-average-gd-filter} and \eqref{eq:proof-projected-bounded-moments} give
\begin{equation}
  \kappa_{1,t}=\eta t(\alpha-\varphi)+O(1),
  \qquad
  \kappa_{2,t}=\eta^2t^2(\alpha-\varphi)+O(1).
  \label{eq:proof-projected-kappa-growth}
\end{equation}
Substituting Equation~\eqref{eq:proof-projected-kappa-growth} into Equation~\eqref{eq:proof-projected-gd-risk} gives
\begin{align}
  \lim_{t\to\infty}\frac{\mathcal E(t)}{t^2}
  &=\frac{\eta^2(1-\varphi)(1-\varphi+\sigma^2)(\alpha-\varphi)}{\alpha^2}
    +\frac{\eta^2(1-\varphi)(\alpha-\varphi)^2}{\alpha^2}\\
  &=\frac{\eta^2(1-\varphi)(\alpha-\varphi)
      (\alpha+1-2\varphi+\sigma^2)}{\alpha^2},
\end{align}
which proves Equation~\eqref{eq:phase-two-gd-risk-growth}.
\end{proof}

The rank deficit produces a unit-eigenvalue sector in the learned iteration.
Each inference step adds a source component in that sector, so the state grows linearly and the squared error grows quadratically.
Together with full-GD recovery for \(\varphi>1\), Proposition~\ref{prop:projected-gd-long-time} completes the \(\alpha>1\) part of Result~\ref{res:long-time-phases}.

\section{The \(0<\alpha<1\) Case}
\label{app:proof-phase-three-long-time}
\label{app:task-cavity}

In this section, we derive the \(\alpha<1\) part of Result~\ref{res:long-time-phases} by the cavity method and combine it with the exact \(\alpha>1\) calculation.

\subsection{Statistical properties of learned matrix}
First, we compute the moments of the learned matrix below the GD-recovery threshold using the cavity method in random matrix theory.

\begin{proposition}[Moments of the learned matrix]
\label{prop:cavity-moments}
Suppose \(0<\alpha<1\), \(0<\varphi<1/\alpha\), and \(\lambda\to0\).
With the large-\(D\) replacements stated explicitly in the proof,
\begin{align}
  -\lim_{D\to\infty}\frac{\operatorname{tr}\hat{\mat A}}{\eta D}
  =&\lim_{D\to\infty}\frac{\|\hat{\mat A}\|_{\mathrm F}^2}{\eta^2D}
   =\frac{\varphi}{1+(1-\alpha)\varphi},\\
  \lim_{D\to\infty}
     \frac{\|\hat{\mat A}+\eta\mat I_D\|_{\mathrm F}^2}{\eta^2D}
  =&\frac{1-\alpha\varphi}{1+(1-\alpha)\varphi},
  \label{eq:cavity-frobenius-error}
\end{align}
and
\begin{equation}
  \lim_{D\to\infty}\frac{\operatorname{tr}(\hat{\mat A}^2)}{\eta^2D}
  =\frac{\varphi[1+(1-2\alpha)\varphi]}
       {[1+(1-\alpha)\varphi]^2(1-\alpha^2\varphi)}.
  \label{eq:proof-cavity-transpose-moment}
\end{equation}
\end{proposition}

\begin{proof}
Proposition~\ref{prop:gd-zero-error} shows that \(\mathcal L_{\mathrm{eff}}(\mat A_\star)=0\) for every \(\alpha\).
Because the training objective is quadratic, its \(\lambda\to0\) limit selects the minimum-norm matrix among those satisfying \(\mathcal L_{\mathrm{eff}}(\mat A)=0\).
The training contexts are then rank deficient, so the constraints need not identify the update.
Define
\begin{align}
  \mat P_\mu:=&\mat X_\mu(\mat X_\mu^\top\mat X_\mu)^{-1}\mat X_\mu^\top,\\
  \mat Q_\mu:=&\mat S_\mu(\mat S_\mu^\top\mat S_\mu)^{-1}\mat S_\mu^\top,\\
  \mat S_\mu:=&(\vb s_{\mu,1},\ldots,\vb s_{\mu,K-1}).
  \label{eq:rank-deficient-projectors}
\end{align}
By Equation~\eqref{eq:rank-deficient-projectors}, the teacher states lie in the context span and \(\mat P_\mu\mat Q_\mu=\mat Q_\mu\).
The condition \(\mathcal L_{\mathrm{eff}}(\mat A)=0\) is equivalent to
\begin{align}
  \mat H_\mu(\mat A+\eta\mat I_D)\mat S_\mu=\mat0
  \iff&\mat P_\mu(\mat A+\eta\mat I_D)\mat Q_\mu=\mat0\\
  \iff&(\mat Q_\mu\otimes\mat P_\mu)
      \operatorname{vec}(\mat A+\eta\mat I_D)=\vb0.
  \label{eq:proof-flag-constraints}
\end{align}
The minimum-norm solution is unchanged if the original loss is replaced by the squared norm of these equivalent constraints.
With an auxiliary regularizer \(\epsilon>0\), define
\begin{equation}
  \mathcal R_\epsilon:=
     \left[\sum_{\mu=1}^M\mat Q_\mu\otimes\mat P_\mu
             +\epsilon\mat I_{D^2}\right]^{-1}.
  \label{eq:cavity-identification-resolvent}
\end{equation}
Vectorizing Equation~\eqref{eq:proof-flag-constraints} and using Equation~\eqref{eq:cavity-identification-resolvent} gives
\begin{align}
  \operatorname{vec}(\hat{\mat A})
  =&-\eta\lim_{\epsilon\to0}
      \mathcal R_\epsilon
      \left[\sum_\mu\mat Q_\mu\otimes\mat P_\mu\right]
      \operatorname{vec}\mat I_D\\
  =&-\eta\operatorname{vec}\mat I_D
      +\eta\lim_{\epsilon\to0}
           \epsilon\mathcal R_\epsilon\operatorname{vec}\mat I_D.
  \label{eq:proof-cavity-ridge-solution}
\end{align}
Equation~\eqref{eq:proof-cavity-ridge-solution} gives
\begin{equation}
  \frac{\operatorname{vec}(\hat{\mat A})}{\eta}
  =-\operatorname{vec}\mat I_D
   +\left[\lim_{\epsilon\to0}\epsilon\mathcal R_\epsilon\right]
      \operatorname{vec}\mat I_D.
\end{equation}
The matrix \(\lim_{\epsilon\to0}\epsilon\mathcal R_\epsilon\) is an orthogonal projector, and hence its transpose and its square are equal to itself.
Therefore,
\begin{align}
 -\frac{\operatorname{tr}\hat{\mat A}}{\eta D}
 =&1-\frac1D(\operatorname{vec}\mat I_D)^\top
 \left[\lim_{\epsilon\to0}\epsilon\mathcal R_\epsilon\right]
 \operatorname{vec}\mat I_D,\\
 \frac{\|\hat{\mat A}\|_{\mathrm F}^2}{\eta^2D}
 =&\frac1D(\operatorname{vec}\mat I_D)^\top
 \left[\mat I_{D^2}-
 \lim_{\epsilon\to0}\epsilon\mathcal R_\epsilon\right]^2
 \operatorname{vec}\mat I_D\\
 =&1-\frac1D(\operatorname{vec}\mat I_D)^\top
 \left[\lim_{\epsilon\to0}\epsilon\mathcal R_\epsilon\right]
 \operatorname{vec}\mat I_D,\\
 \frac{\operatorname{tr}(\hat{\mat A}^2)}{\eta^2D}
 =&1-\frac2D(\operatorname{vec}\mat I_D)^\top
 \left[\lim_{\epsilon\to0}\epsilon\mathcal R_\epsilon\right]
 \operatorname{vec}\mat I_D\\
 &+\frac1D(\operatorname{vec}\mat I_D)^\top
 \left[\lim_{\epsilon\to0}\epsilon\mathcal R_\epsilon\right]
 \mathrm T_D
 \left[\lim_{\epsilon\to0}\epsilon\mathcal R_\epsilon\right]
 \operatorname{vec}\mat I_D,
 \label{eq:proof-cavity-trace-target}
\end{align}
where \(\mathrm T_D:\R^{D^2}\to\R^{D^2}\) denotes the transpose operator defined by
\begin{equation}
   \mathrm T_D\operatorname{vec}(\mat B)
   =\operatorname{vec}(\mat B^\top).
\end{equation}

It therefore suffices to calculate the following two scalars separately:
\begin{equation}
 \frac1D(\operatorname{vec}\mat I_D)^\top
 \left[\lim_{\epsilon\to0}\epsilon\mathcal R_\epsilon\right]
 \operatorname{vec}\mat I_D,
 \label{eq:proof-cavity-first-targets}
\end{equation}
and
\begin{equation}
 \frac1D(\operatorname{vec}\mat I_D)^\top
 \left[\lim_{\epsilon\to0}\epsilon\mathcal R_\epsilon\right]
 \mathrm T_D
 \left[\lim_{\epsilon\to0}\epsilon\mathcal R_\epsilon\right]
 \operatorname{vec}\mat I_D.
 \label{eq:proof-cavity-second-targets}
\end{equation}

\paragraph{Calculation of Equation~\eqref{eq:proof-cavity-first-targets}.}
First, we calculate Equation~\eqref{eq:proof-cavity-first-targets}.
Define the order parameter:
\begin{equation}
  \chi:=-\lim_{D\to\infty}
      \frac{\operatorname{tr}\hat{\mat A}}{\eta D}.
\end{equation}
Equations~\eqref{eq:proof-cavity-trace-target} and \eqref{eq:proof-cavity-first-targets} give
\begin{equation}
  \frac1D(\operatorname{vec}\mat I_D)^\top
  \left[\lim_{\epsilon\to0}
       \epsilon\mathcal R_\epsilon\right]
  \operatorname{vec}\mat I_D
  =1-\chi.
  \label{eq:proof-cavity-identity-overlap}
\end{equation}

Since \(\mathcal R_\epsilon\) is the resolvent of a sum of independent task contributions, we use a leave-one-task cavity calculation, in which the removed task is independent of the resolvent formed by the remaining tasks.
In this approach, deleting task \(\mu\) defines
\begin{equation}
  \mathcal R_\epsilon^{-\mu}
  :=\left[
      \sum_{\nu\ne\mu}\mat Q_\nu\otimes\mat P_\nu
      +\epsilon\mat I_{D^2}
    \right]^{-1}.
  \label{eq:proof-cavity-leave-one-task-resolvent}
\end{equation}
Because \(\mat P_\mu\) and \(\mat Q_\mu\) are projectors,
\begin{align}
 &(\mat Q_\mu\otimes\mat P_\mu)^2
 =(\mat Q_\mu^2\otimes\mat P_\mu^2)
  =\mat Q_\mu\otimes\mat P_\mu\\
 &(\operatorname{vec}\mat I_D)^\top
 (\mat Q_\mu\otimes\mat P_\mu)
 \operatorname{vec}\mat I_D
 =\operatorname{tr}(\mat P_\mu\mat Q_\mu)
  =K-1.
 \label{eq:proof-cavity-projector-source}
\end{align}
Applying the Woodbury formula to Equations~\eqref{eq:cavity-identification-resolvent} and \eqref{eq:proof-cavity-leave-one-task-resolvent}, and using Equation~\eqref{eq:proof-cavity-projector-source}, gives
\begin{align}
 (\mat Q_\mu\otimes\mat P_\mu)\mathcal R_\epsilon
 =\left[\epsilon\mat I_{D^2}
  +(\mat Q_\mu\otimes\mat P_\mu)
   \epsilon\mathcal R_\epsilon^{-\mu}
   (\mat Q_\mu\otimes\mat P_\mu)\right]^{-1}
 (\mat Q_\mu\otimes\mat P_\mu)
   \epsilon\mathcal R_\epsilon^{-\mu}.
 \label{eq:proof-cavity-tested-resolvent}
\end{align}
We first evaluate the matrix inside the inverse in Equation~\eqref{eq:proof-cavity-tested-resolvent}.
The matrix \(\mathcal R_\epsilon^{-\mu}\) depends only on the tasks \(\nu\ne\mu\), and is therefore independent of \((\mat P_\mu,\mat Q_\mu)\).
Throughout this cavity calculation, \(\simeq\) denotes equality at the retained
leading order after the displayed normalization, with discarded terms
vanishing in probability as \(D\to\infty\).
Its compression by task \(\mu\) concentrates around its conditional average, so
\begin{align}
 (\mat Q_\mu\otimes\mat P_\mu)
   \left[\lim_{\epsilon\to0}\epsilon
      \mathcal R_\epsilon^{-\mu}\right]
   (\mat Q_\mu\otimes\mat P_\mu)
 \simeq&
   \frac1{D^2}\operatorname{tr}\left[
      \lim_{\epsilon\to0}\epsilon\mathcal R_\epsilon^{-\mu}
   \right]
   (\mat Q_\mu\otimes\mat P_\mu).
\end{align}
To calculate the normalized trace in this expression, let \(\lambda_1,\ldots,\lambda_{D^2}\) be the eigenvalues of
\begin{equation}
 \sum_{\nu\ne\mu}\mat Q_\nu\otimes\mat P_\nu.
\end{equation}
The eigenvalues of \(\mathcal R_\epsilon^{-\mu}\) are \(1/(\lambda_j+\epsilon)\), and hence the eigenvalues of \(\epsilon\mathcal R_\epsilon^{-\mu}\) satisfy
\begin{equation}
 \frac{\epsilon}{\lambda_j+\epsilon}
 \to
 \begin{cases}
  0,&\lambda_j>0,\\
  1,&\lambda_j=0.
 \end{cases}
\end{equation}
Therefore,
\begin{align}
 \operatorname{tr}\left[
  \lim_{\epsilon\to0}\epsilon\mathcal R_\epsilon^{-\mu}
 \right]
 =\#\{j:\lambda_j=0\}
 =D^2-\operatorname{rank}\left(
   \sum_{\nu\ne\mu}\mat Q_\nu\otimes\mat P_\nu
 \right).
\end{align}
For each task,
\begin{align}
 \operatorname{rank}(\mat Q_\nu\otimes\mat P_\nu)
 =&\operatorname{rank}(\mat Q_\nu)
   \operatorname{rank}(\mat P_\nu)
  =L(K-1).
\end{align}
Since the constraints from different tasks are linearly independent, 
\begin{equation}
 \operatorname{rank}\left(
  \sum_{\nu\ne\mu}\mat Q_\nu\otimes\mat P_\nu
 \right)
 =(M-1)L(K-1).
\end{equation}
It follows that
\begin{align}
 \frac1{D^2}\operatorname{tr}\left[
  \lim_{\epsilon\to0}\epsilon\mathcal R_\epsilon^{-\mu}
 \right]
 =&1-\frac{(M-1)L(K-1)}{D^2}\\
 =&1-\frac LD\frac{(M-1)(K-1)}D\\
 \to&1-\alpha\varphi.
 \label{eq:proof-cavity-zero-eigenvalue-fraction}
\end{align}
Consequently, the compression in Equation~\eqref{eq:proof-cavity-tested-resolvent} satisfies
\begin{align}
 (\mat Q_\mu\otimes\mat P_\mu)
   \left[\lim_{\epsilon\to0}\epsilon
      \mathcal R_\epsilon^{-\mu}\right]
   (\mat Q_\mu\otimes\mat P_\mu)
 \simeq(1-\alpha\varphi)
   (\mat Q_\mu\otimes\mat P_\mu).
 \label{eq:proof-cavity-projector-compression}
\end{align}
Multiplying Equation~\eqref{eq:proof-cavity-tested-resolvent} by \(\operatorname{vec}\mat I_D\) and then by \((\operatorname{vec}\mat I_D)^\top/(K-1)\) from the left gives
\begin{align}
 &\lim_{\epsilon\to0}\frac1{K-1}
   (\operatorname{vec}\mat I_D)^\top
   (\mat Q_\mu\otimes\mat P_\mu)
   \mathcal R_\epsilon\operatorname{vec}\mat I_D\\
 =&\lim_{\epsilon\to0}\frac1{K-1}
   (\operatorname{vec}\mat I_D)^\top
   \left[\epsilon\mat I_{D^2}
    +(\mat Q_\mu\otimes\mat P_\mu)
      \epsilon\mathcal R_\epsilon^{-\mu}
      (\mat Q_\mu\otimes\mat P_\mu)\right]^{-1}(\mat Q_\mu\otimes\mat P_\mu)
   \epsilon\mathcal R_\epsilon^{-\mu}
   \operatorname{vec}\mat I_D\\
 \simeq&\lim_{\epsilon\to0}\frac1{K-1}
   (\operatorname{vec}\mat I_D)^\top
   \left[\epsilon\mat I_{D^2}
    +(1-\alpha\varphi)(\mat Q_\mu\otimes\mat P_\mu)\right]^{-1}
   (\mat Q_\mu\otimes\mat P_\mu)
   \epsilon\mathcal R_\epsilon^{-\mu}
   \operatorname{vec}\mat I_D\\
 =&\lim_{\epsilon\to0}
   \frac1{(K-1)(\epsilon+1-\alpha\varphi)}
   (\operatorname{vec}\mat I_D)^\top
   (\mat Q_\mu\otimes\mat P_\mu)
   \epsilon\mathcal R_\epsilon^{-\mu}
   \operatorname{vec}\mat I_D\\
 =&\frac1{(K-1)(1-\alpha\varphi)}
   (\operatorname{vec}\mat I_D)^\top
   (\mat Q_\mu\otimes\mat P_\mu)
   \left[\lim_{\epsilon\to0}\epsilon
      \mathcal R_\epsilon^{-\mu}\right]
   \operatorname{vec}\mat I_D\\
 \simeq&\frac1{1-\alpha\varphi}
   \frac1D(\operatorname{vec}\mat I_D)^\top
   \left[\lim_{\epsilon\to0}\epsilon
      \mathcal R_\epsilon^{-\mu}\right]
   \operatorname{vec}\mat I_D\\
 \simeq&\frac1{1-\alpha\varphi}
   \frac1D(\operatorname{vec}\mat I_D)^\top
   \left[\lim_{\epsilon\to0}\epsilon
      \mathcal R_\epsilon\right]
   \operatorname{vec}\mat I_D\\
 =&\frac{1-\chi}{1-\alpha\varphi}.
 \label{eq:proof-cavity-first-target-right}
\end{align}
Here the first \(\simeq\) uses Equation~\eqref{eq:proof-cavity-projector-compression}.
The following equality uses
\begin{align}
 &\left[\epsilon\mat I_{D^2}
   +(1-\alpha\varphi)(\mat Q_\mu\otimes\mat P_\mu)\right]^{-1}
   (\mat Q_\mu\otimes\mat P_\mu)
 =\frac1{\epsilon+1-\alpha\varphi}
   (\mat Q_\mu\otimes\mat P_\mu),
\end{align}
which follows from \((\mat Q_\mu\otimes\mat P_\mu)^2=\mat Q_\mu\otimes\mat P_\mu\).
The second \(\simeq\) uses the following large-\(D\) concentration for the contraction with the independent task \(\mu\):
\begin{align}
 \frac1{K-1}(\operatorname{vec}\mat I_D)^\top
   (\mat Q_\mu\otimes\mat P_\mu)
   \left[\lim_{\epsilon\to0}\epsilon
      \mathcal R_\epsilon^{-\mu}\right]
   \operatorname{vec}\mat I_D
 \simeq\frac1D(\operatorname{vec}\mat I_D)^\top
   \left[\lim_{\epsilon\to0}
      \epsilon\mathcal R_\epsilon^{-\mu}\right]
   \operatorname{vec}\mat I_D.
 \label{eq:proof-cavity-source-contraction}
\end{align}

On the other hand, averaging the left-hand side over \(\mu\) gives
\begin{align}
 \frac1{M(K-1)}(\operatorname{vec}\mat I_D)^\top
   \left[\sum_{\mu=1}^M\mat Q_\mu\otimes\mat P_\mu\right]
   \mathcal R_\epsilon\operatorname{vec}\mat I_D
 =\frac1{M(K-1)}(\operatorname{vec}\mat I_D)^\top
    (\mat I_{D^2}-\epsilon\mathcal R_\epsilon)
    \operatorname{vec}\mat I_D
  \to\frac{\chi}{\varphi}.
  \label{eq:proof-cavity-first-target-left}
\end{align}
Equating Equations~\eqref{eq:proof-cavity-first-target-right} and \eqref{eq:proof-cavity-first-target-left} gives the following self-consistent equation:
\begin{align}
  \frac{\chi}{\varphi}
  =&\frac{1-\chi}{1-\alpha\varphi},
  \label{eq:proof-cavity-scalar-signal}
\end{align}
which gives
\begin{align}
  \chi=&\frac{\varphi}{1+(1-\alpha)\varphi}.
\end{align}

The estimator is the orthogonal projection of \(-\eta\mat I_D\) onto the row space of the constraints.
Therefore
\begin{align}
  \lim_{D\to\infty}\frac{\|\hat{\mat A}\|_{\mathrm F}^2}{\eta^2D}
  =&-\lim_{D\to\infty}\frac{\operatorname{tr}\hat{\mat A}}{\eta D}
    =\chi
    =\frac{\varphi}{1+(1-\alpha)\varphi}.
\end{align}

Equation~\eqref{eq:proof-cavity-ridge-solution} also gives
\begin{align}
  \lim_{D\to\infty}
  \frac{\|\hat{\mat A}+\eta\mat I_D\|_{\mathrm F}^2}{\eta^2D}
  =&\lim_{D\to\infty}\frac1D
  (\operatorname{vec}\mat I_D)^\top
  \left[\lim_{\epsilon\to0}\epsilon\mathcal R_\epsilon\right]^2
  \operatorname{vec}\mat I_D\\
  =&\lim_{D\to\infty}\frac1D
  (\operatorname{vec}\mat I_D)^\top
  \left[\lim_{\epsilon\to0}\epsilon\mathcal R_\epsilon\right]
  \operatorname{vec}\mat I_D\\
  =&1-\chi
  =\frac{1-\alpha\varphi}{1+(1-\alpha)\varphi}.
\end{align}
The second equality uses the projector identity
\(\left[\lim_{\epsilon\to0}\epsilon\mathcal R_\epsilon\right]^2
=\lim_{\epsilon\to0}\epsilon\mathcal R_\epsilon\).
The third equality uses Equation~\eqref{eq:proof-cavity-identity-overlap}.

\paragraph{Calculation of Equation~\eqref{eq:proof-cavity-second-targets}.}
Second, we calculate Equation~\eqref{eq:proof-cavity-second-targets} by deriving a second self-consistent equation from the same Woodbury identity.
Define the order parameters:
\begin{align}
  m_{\mathrm F}
  :=&\lim_{D\to\infty}\frac1D
      \|\hat{\mat A}+\eta\chi\mat I_D\|_{\mathrm F}^2,\\
  m_{\mathrm T}
  :=&\lim_{D\to\infty}\frac1D
      \operatorname{tr}(\hat{\mat A}+\eta\chi\mat I_D)^2.
\end{align}
By Equations~\eqref{eq:proof-cavity-ridge-solution} and \eqref{eq:proof-cavity-identity-overlap}, the second target satisfies
\begin{align}
 \frac{m_{\mathrm T}}{\eta^2}
 =&\lim_{D\to\infty}\left\{
 \frac1D(\operatorname{vec}\mat I_D)^\top
 \left[\lim_{\epsilon\to0}\epsilon\mathcal R_\epsilon\right]
 \mathrm T_D
 \left[\lim_{\epsilon\to0}\epsilon\mathcal R_\epsilon\right]
 \operatorname{vec}\mat I_D
 -(1-\chi)^2\right\}.
 \label{eq:proof-cavity-second-target-centered}
\end{align}
The trace and Frobenius identities obtained above give
\begin{align}
 m_{\mathrm F}
 =&\lim_{D\to\infty}\left[
   \frac{\|\hat{\mat A}\|_{\mathrm F}^2}{D}
   +\frac{2\eta\chi}{D}\operatorname{tr}\hat{\mat A}
   +\eta^2\chi^2\right]\\
 =&\eta^2\chi-2\eta^2\chi^2+\eta^2\chi^2\\
 =&\eta^2(\chi-\chi^2).
\end{align}
We now start again from the Woodbury identity (Equation~\eqref{eq:proof-cavity-tested-resolvent}):
\begin{align}
 (\mat Q_\mu\otimes\mat P_\mu)
   \mathcal R_\epsilon\operatorname{vec}\mat I_D
 =\left[\epsilon\mat I_{D^2}
  +(\mat Q_\mu\otimes\mat P_\mu)
   \epsilon\mathcal R_\epsilon^{-\mu}
   (\mat Q_\mu\otimes\mat P_\mu)\right]^{-1}
 (\mat Q_\mu\otimes\mat P_\mu)
   \epsilon\mathcal R_\epsilon^{-\mu}
   \operatorname{vec}\mat I_D.
 \label{eq:proof-cavity-second-woodbury}
\end{align}
Equation~\eqref{eq:proof-cavity-ridge-solution} can also be written as
\begin{align}
 \frac{\operatorname{vec}(\hat{\mat A}+\eta\chi\mat I_D)}{\eta}
 =&\chi\operatorname{vec}\mat I_D
 -\lim_{\epsilon\to0}\sum_{\mu=1}^M
  (\mat Q_\mu\otimes\mat P_\mu)
  \mathcal R_\epsilon\operatorname{vec}\mat I_D.
 \label{eq:proof-cavity-centered-task-sum}
\end{align}
Applying \(\eta^2D^{-1}(\cdot)^\top\mathrm T_D(\cdot)\) to both sides of Equation~\eqref{eq:proof-cavity-centered-task-sum} gives
\begin{align}
 &\lim_{D\to\infty}\frac1D
  \operatorname{vec}(\hat{\mat A}+\eta\chi\mat I_D)^\top
  \mathrm T_D
  \operatorname{vec}(\hat{\mat A}+\eta\chi\mat I_D)\\
 =&\lim_{D\to\infty}\frac{\eta^2}{D}
  \left[\chi\operatorname{vec}\mat I_D
  -\lim_{\epsilon\to0}\sum_{\mu=1}^M
  (\mat Q_\mu\otimes\mat P_\mu)
  \mathcal R_\epsilon\operatorname{vec}\mat I_D\right]^\top
  \mathrm T_D\notag\\
 &\quad\times
  \left[\chi\operatorname{vec}\mat I_D
  -\lim_{\epsilon\to0}\sum_{\nu=1}^M
  (\mat Q_\nu\otimes\mat P_\nu)
  \mathcal R_\epsilon\operatorname{vec}\mat I_D\right]\\
 =&\lim_{D\to\infty}\frac{\eta^2}{D}\Bigg\{
  \chi^2(\operatorname{vec}\mat I_D)^\top
  \mathrm T_D\operatorname{vec}\mat I_D
  -2\chi\lim_{\epsilon\to0}\sum_{\mu=1}^M
  (\operatorname{vec}\mat I_D)^\top\mathrm T_D
  (\mat Q_\mu\otimes\mat P_\mu)
  \mathcal R_\epsilon\operatorname{vec}\mat I_D\notag\\
 &\qquad+\lim_{\epsilon\to0}\sum_{\mu,\nu=1}^M
  (\operatorname{vec}\mat I_D)^\top
  \mathcal R_\epsilon(\mat Q_\mu\otimes\mat P_\mu)
  \mathrm T_D(\mat Q_\nu\otimes\mat P_\nu)
  \mathcal R_\epsilon\operatorname{vec}\mat I_D\Bigg\}\\
 =&\lim_{D\to\infty}\frac{\eta^2}{D}\Bigg\{
  \chi^2(\operatorname{vec}\mat I_D)^\top
  \mathrm T_D\operatorname{vec}\mat I_D
  -2\chi\lim_{\epsilon\to0}\sum_{\mu=1}^M
  (\operatorname{vec}\mat I_D)^\top\mathrm T_D
  (\mat Q_\mu\otimes\mat P_\mu)
  \mathcal R_\epsilon\operatorname{vec}\mat I_D\notag\\
 &\qquad+\lim_{\epsilon\to0}\sum_{\mu=\nu}
  (\operatorname{vec}\mat I_D)^\top
  \mathcal R_\epsilon(\mat Q_\mu\otimes\mat P_\mu)
  \mathrm T_D(\mat Q_\nu\otimes\mat P_\nu)
  \mathcal R_\epsilon\operatorname{vec}\mat I_D\notag\\
 &\qquad+\lim_{\epsilon\to0}\sum_{\mu\ne\nu}
  (\operatorname{vec}\mat I_D)^\top
  \mathcal R_\epsilon(\mat Q_\mu\otimes\mat P_\mu)
  \mathrm T_D(\mat Q_\nu\otimes\mat P_\nu)
  \mathcal R_\epsilon\operatorname{vec}\mat I_D\Bigg\}.
 \label{eq:proof-cavity-second-quadratic-identity}
\end{align}
The left-hand side of Equation~\eqref{eq:proof-cavity-second-quadratic-identity} is
\begin{align}
 &\lim_{D\to\infty}\frac1D
  \operatorname{vec}(\hat{\mat A}+\eta\chi\mat I_D)^\top
  \mathrm T_D
  \operatorname{vec}(\hat{\mat A}+\eta\chi\mat I_D)\\
 =&\lim_{D\to\infty}\frac1D
  \operatorname{tr}(\hat{\mat A}+\eta\chi\mat I_D)^2\\
 =&m_{\mathrm T}.
\end{align}
For each task contribution on the right-hand side of Equation~\eqref{eq:proof-cavity-second-quadratic-identity}, we substitute Equation~\eqref{eq:proof-cavity-second-woodbury}.
Next, to calculate the right-hand side, we introduce the following proposition.

\begin{proposition}[Moments of the task projectors]
\label{prop:task-projector-moments}
Let \(\mat B\in\R^{D\times D}\) be any matrix independent of task \(\mu\).
\begin{align}
 \E_\mu[\mat P_\mu\mat B\mat P_\mu]
 \simeq&\alpha^2\mat B+
 \frac{\alpha(1-\alpha)}D
 \left(\mat B^\top+\operatorname{tr}(\mat B)\mat I_D\right),\\
 \E_\mu[\mat Q_\mu\mat B\mat Q_\mu]
 \simeq&\frac{(K-1)^2}{D^2}\mat B+
 \frac{K-1}{D^2}
 \left(\mat B^\top+\operatorname{tr}(\mat B)\mat I_D\right).
 \label{eq:task-projector-second-moments}
\end{align}
\end{proposition}

\begin{proof}
Marginally, \(\mat P_\mu\) and \(\mat Q_\mu\) are rotationally invariant orthogonal projectors of ranks \(L\) and \(K-1\), respectively.
The entrywise second moment of either projector has the form
\begin{align}
 \E_\mu[(\mat\Pi_\mu)_{ij}(\mat\Pi_\mu)_{kl}]
 =&a_{D,r}\delta_{ij}\delta_{kl}
 +b_{D,r}(\delta_{ik}\delta_{jl}+\delta_{il}\delta_{jk}),
\end{align}
where \(\mat\Pi_\mu=\mat P_\mu\) for \(r=L\) and \(\mat\Pi_\mu=\mat Q_\mu\) for \(r=K-1\).
Using \((\operatorname{tr}\mat\Pi_\mu)^2=r^2\) and \(\operatorname{tr}(\mat\Pi_\mu^2)=r\) gives
\begin{align}
 D^2a_{D,r}+2Db_{D,r}
 =&r^2,\\
 Da_{D,r}+(D^2+D)b_{D,r}
 =&r.
\end{align}
Solving these equations gives
\begin{align}
 a_{D,r}
 =&\frac{r[r(D+1)-2]}{D(D-1)(D+2)},\\
 b_{D,r}
 =&\frac{r(D-r)}{D(D-1)(D+2)}.
\end{align}
Therefore,
\begin{align}
 \E_\mu[(\mat\Pi_\mu\mat B\mat\Pi_\mu)_{il}]
 =&\sum_{j,k}(\mat B)_{jk}
 \E_\mu[(\mat\Pi_\mu)_{ij}(\mat\Pi_\mu)_{kl}]\\
 =&a_{D,r}\mat B_{il}
 +b_{D,r}\mat B_{li}
 +b_{D,r}\operatorname{tr}(\mat B)\delta_{il}.
\end{align}
Substituting \(r=L=\alpha D\) and \(r=K-1\), and expanding the coefficients at their retained large-\(D\) orders, gives Equation~\eqref{eq:task-projector-second-moments}.
\end{proof}

After substituting Equation~\eqref{eq:proof-cavity-second-woodbury}, we use concentration of the normalized contractions around their conditional task averages.
For \(\mu\ne\nu\), independence of the two tasks gives
\begin{align}
  \E\operatorname{tr}(\mat P_\mu\mat P_\nu)
  =&\operatorname{tr}\left(
      \E\mat P_\mu\E\mat P_\nu\right)
   =\operatorname{tr}\left(
      \frac LD\mat I_D\frac LD\mat I_D\right)
   =\frac{L^2}{D},\\
  \E\operatorname{tr}(\mat Q_\mu\mat Q_\nu)
  =&\operatorname{tr}\left(
      \E\mat Q_\mu\E\mat Q_\nu\right)
   =\operatorname{tr}\left(
      \frac{K-1}{D}\mat I_D
      \frac{K-1}{D}\mat I_D\right)
   =\frac{(K-1)^2}{D}.
\end{align}
We now calculate the two cases separately.
For \(\mu=\nu\), Equation~\eqref{eq:task-projector-second-moments} gives
\begin{align}
 &\left\{\frac1{D^2}\operatorname{tr}\left[
   \lim_{\epsilon\to0}\epsilon\mathcal R_\epsilon^{-\mu}
   \right]\right\}m_{\mathrm F}
 =\left[1-\frac{(M-1)L(K-1)}{D^2}\right]m_{\mathrm F}\\
 \to&(1-\alpha\varphi)m_{\mathrm F}.
\end{align}
For \(\mu\ne\nu\), independence of the two tasks gives
\begin{align}
 &\sum_{\nu\ne\mu}\frac1{K-1}
   \E\operatorname{tr}(\mat Q_\mu\mat Q_\nu)
   \frac1D\E\operatorname{tr}(\mat P_\mu\mat P_\nu)
   m_{\mathrm T}\\
 =&(M-1)\frac1{K-1}\frac{(K-1)^2}{D}
   \frac1D\frac{L^2}{D}m_{\mathrm T}\\
 =&(M-1)\frac{K-1}{D}\frac{L^2}{D^2}m_{\mathrm T}\\
 \to&\alpha^2\varphi m_{\mathrm T}.
\end{align}
Equating the two sides of Equation~\eqref{eq:proof-cavity-second-quadratic-identity} gives
\begin{align}
  m_{\mathrm T}
  =&\lim_{D\to\infty}
    \left[1-\frac{(M-1)L(K-1)}{D^2}\right]m_{\mathrm F}
  +\lim_{D\to\infty}
    (M-1)\frac{K-1}{D}\frac{L^2}{D^2}m_{\mathrm T}
    \\
  =&(1-\alpha\varphi)m_{\mathrm F}
    +\alpha^2\varphi m_{\mathrm T},
  \label{eq:proof-cavity-second-self-consistency}
\end{align}
which gives
\begin{align}
  \frac{m_{\mathrm T}}{\eta^2}
  =&\frac{\varphi(1-\alpha\varphi)^2}
    {[1+(1-\alpha)\varphi]^2(1-\alpha^2\varphi)}.
\end{align}
Since the centered matrix has zero normalized trace,
\begin{align}
  \lim_{D\to\infty}\frac{\operatorname{tr}(\hat{\mat A}^2)}{\eta^2D}
  =\chi^2+\frac{m_{\mathrm T}}{\eta^2}
  =\frac{\varphi[1+(1-2\alpha)\varphi]}
       {[1+(1-\alpha)\varphi]^2(1-\alpha^2\varphi)}.
\end{align}
This proves Equation~\eqref{eq:proof-cavity-transpose-moment}.
\end{proof}

\subsection{Moment-matched Gaussian effective matrix}

Proposition~\ref{prop:cavity-moments} determines the normalized trace, Frobenius norm, and transpose moment of \(\hat{\mat A}\).
Directly analyzing the resolvent of \(\hat{\mat A}\) is difficult, so, following \citet{takanami2025lowrank}, we introduce a tractable Gaussian surrogate for \(\hat{\mat A}\).
This replacement is an assumption, but we expect it to become accurate in the high-dimensional limit.
Specifically, the surrogate matrix is chosen to reproduce the normalized trace, Frobenius norm, and transpose moment in Proposition~\ref{prop:cavity-moments}.

Let
\begin{equation}
  r=\operatorname{rank}\left(\sum_{\mu=1}^M\mat Q_\mu\right)
  =\min\{D,M(K-1)\}.
\end{equation}

Define \(\mat O_\parallel\in\mathbb R^{D\times r}\) as the matrix whose columns are orthonormal eigenvectors of \(\sum_{\mu=1}^M\mat Q_\mu\) corresponding to its nonzero eigenvalues.
Define \(\mat O_\perp\in\mathbb R^{D\times(D-r)}\) as an orthonormal basis of the null space of \(\sum_{\mu=1}^M\mat Q_\mu\).
Then we have

\begin{equation}
  \sum_{\mu=1}^M\mat Q_\mu
  =\mat O_\parallel\operatorname{diag}(x_1,\ldots,x_r)\mat O_\parallel^\top,
  \qquad
  x_1,\ldots,x_r>0,
  \qquad
  (\mat O_\parallel,\mat O_\perp)^\top
  (\mat O_\parallel,\mat O_\perp)=\mat I_D.
  \label{eq:proof-state-projector-eigendecomposition}
\end{equation}

\begin{proposition}[Moment-matched effective expression]
\label{prop:cavity-matrix-profile}
Suppose \(0<\alpha<1\), \(0<\varphi<1/\alpha\), and \(\lambda\to0\).
Define \(\hat{\mat A}_{\mathrm{eff}}\) by
\begin{align}
  \frac{\hat{\mat A}_{\mathrm{eff}}\mat O_\parallel}{\eta}
  =&-\mat O_\parallel\operatorname{diag}_j
          \frac{x_j}{u_\varphi+v_\varphi x_j}
       +\frac{\mat W}{\sqrt D}\operatorname{diag}_j
          \frac{\sqrt{c_\varphi x_j}}{u_\varphi+v_\varphi x_j},\\
  \hat{\mat A}_{\mathrm{eff}}\mat O_\perp=&\mat0,
  \label{eq:proof-cavity-matrix-profile}
\end{align}
where \(\mat W\in\mathbb R^{D\times r}\) has independent \(N(0,1)\) entries, and
\begin{align}
  u_\varphi
  =&(1-\alpha)(1-\alpha\varphi)[1+(1-\alpha)\varphi],\\
  v_\varphi
  =&\alpha[1+(1-\alpha)\varphi],\\
  c_\varphi
  =&\alpha(1-\alpha)\varphi(1-\alpha\varphi).
  \label{eq:phase-three-matrix-coefficients}
\end{align}
Then
\begin{align}
 -\lim_{D\to\infty}
   \frac{\operatorname{tr}\hat{\mat A}_{\mathrm{eff}}}{\eta D}
 =&\lim_{D\to\infty}
   \frac{\|\hat{\mat A}_{\mathrm{eff}}\|_{\mathrm F}^2}{\eta^2D}
  =\frac{\varphi}{1+(1-\alpha)\varphi},\\
 \lim_{D\to\infty}
   \frac{\operatorname{tr}(\hat{\mat A}_{\mathrm{eff}}^2)}{\eta^2D}
 =&\frac{\varphi[1+(1-2\alpha)\varphi]}
 {[1+(1-\alpha)\varphi]^2(1-\alpha^2\varphi)}.
\end{align}
\end{proposition}

\begin{proof}
For fixed \(K\), the nonzero eigenvalues \(x_j\) of \(\sum_\mu\mat Q_\mu\) obey the Marchenko--Pastur law \(\nu_{1/\varphi}\) defined in Equation~\eqref{eq:mp-average-gd-filter}.
Thus, for \(f(0)=0\),
\begin{equation}
  \frac1D\sum_{j=1}^r f(x_j)
  \longrightarrow\varphi\langle f\rangle_{1/\varphi}.
  \label{eq:proof-state-gram-law}
\end{equation}
For \(\varphi>1\), \(\nu_{1/\varphi}\) has a zero atom of mass \(1-1/\varphi\).

Since \(\hat{\mat A}_{\mathrm{eff}}\mat O_\perp=\mat0\), Equation~\eqref{eq:proof-cavity-matrix-profile} gives
\begin{align}
  \frac{\hat{\mat A}_{\mathrm{eff}}}{\eta}
  =&-\mat O_\parallel\operatorname{diag}_j
       \frac{x_j}{u_\varphi+v_\varphi x_j}\mat O_\parallel^\top
     +\frac{\mat W}{\sqrt D}\operatorname{diag}_j
       \frac{\sqrt{c_\varphi x_j}}{u_\varphi+v_\varphi x_j}
       \mat O_\parallel^\top.
\end{align}
Therefore, the normalized trace satisfies
\begin{align}
 -\frac{\operatorname{tr}\hat{\mat A}_{\mathrm{eff}}}{\eta D}
 =&\frac1D\sum_{j=1}^r
      \frac{x_j}{u_\varphi+v_\varphi x_j}
   -\frac1{D\sqrt D}\operatorname{tr}\left[
      \mat W\operatorname{diag}_j
      \frac{\sqrt{c_\varphi x_j}}{u_\varphi+v_\varphi x_j}
      \mat O_\parallel^\top\right]\\
 \to&\varphi\left\langle
      \frac{x}{u_\varphi+v_\varphi x}
      \right\rangle_{1/\varphi}.
 \label{eq:proof-effective-trace-direct}
\end{align}
Expanding the trace of the square gives
\begin{align}
 \frac{\operatorname{tr}(\hat{\mat A}_{\mathrm{eff}}^2)}{\eta^2D}
 =&\frac1D\operatorname{tr}\left[
      \left(-\mat O_\parallel\operatorname{diag}_j
      \frac{x_j}{u_\varphi+v_\varphi x_j}\mat O_\parallel^\top
      +\frac{\mat W}{\sqrt D}\operatorname{diag}_j
      \frac{\sqrt{c_\varphi x_j}}{u_\varphi+v_\varphi x_j}
      \mat O_\parallel^\top\right)^2\right]
   \nonumber\\
 =&\frac1D\sum_{j=1}^r
      \frac{x_j^2}{(u_\varphi+v_\varphi x_j)^2}
   -\frac{2}{D\sqrt D}\operatorname{tr}\left[
      \mat O_\parallel\operatorname{diag}_j
      \frac{x_j}{u_\varphi+v_\varphi x_j}
      \mat O_\parallel^\top
      \mat W\operatorname{diag}_j
      \frac{\sqrt{c_\varphi x_j}}{u_\varphi+v_\varphi x_j}
      \mat O_\parallel^\top\right]\\
 &+\frac1{D^2}\operatorname{tr}\left[
      \mat W\operatorname{diag}_j
      \frac{\sqrt{c_\varphi x_j}}{u_\varphi+v_\varphi x_j}
      \mat O_\parallel^\top
      \mat W\operatorname{diag}_j
      \frac{\sqrt{c_\varphi x_j}}{u_\varphi+v_\varphi x_j}
      \mat O_\parallel^\top\right]\\
 =&\frac1D\sum_{j=1}^r
      \frac{x_j^2}{(u_\varphi+v_\varphi x_j)^2}
      +o_{\mathbb P}(1)
   \nonumber\\
 \to&\varphi\left\langle
      \frac{x^2}{(u_\varphi+v_\varphi x)^2}
      \right\rangle_{1/\varphi}.
 \label{eq:proof-effective-square-direct}
\end{align}
Similarly, expanding the Frobenius norm gives
\begin{align}
 \frac{\|\hat{\mat A}_{\mathrm{eff}}\|_{\mathrm F}^2}{\eta^2D}
 =&\frac1D\operatorname{tr}\left[
      \begin{aligned}
      &\left(-\mat O_\parallel\operatorname{diag}_j
      \frac{x_j}{u_\varphi+v_\varphi x_j}\mat O_\parallel^\top
      +\frac{\mat W}{\sqrt D}\operatorname{diag}_j
      \frac{\sqrt{c_\varphi x_j}}{u_\varphi+v_\varphi x_j}
      \mat O_\parallel^\top\right)^\top\\
      &\quad\times
      \left(-\mat O_\parallel\operatorname{diag}_j
      \frac{x_j}{u_\varphi+v_\varphi x_j}\mat O_\parallel^\top
      +\frac{\mat W}{\sqrt D}\operatorname{diag}_j
      \frac{\sqrt{c_\varphi x_j}}{u_\varphi+v_\varphi x_j}
      \mat O_\parallel^\top\right)
      \end{aligned}\right]
   \nonumber\\
 =&\frac1D\sum_{j=1}^r
      \frac{x_j^2}{(u_\varphi+v_\varphi x_j)^2}
   -\frac{2}{D\sqrt D}\operatorname{tr}\left[
      \operatorname{diag}_j
      \frac{x_j}{u_\varphi+v_\varphi x_j}
      \mat O_\parallel^\top\mat W
      \operatorname{diag}_j
      \frac{\sqrt{c_\varphi x_j}}{u_\varphi+v_\varphi x_j}\right]\\
 &+\frac1{D^2}\operatorname{tr}\left[
      \operatorname{diag}_j
      \frac{\sqrt{c_\varphi x_j}}{u_\varphi+v_\varphi x_j}
      \mat W^\top\mat W
      \operatorname{diag}_j
      \frac{\sqrt{c_\varphi x_j}}{u_\varphi+v_\varphi x_j}\right]\\
 =&\frac1D\sum_{j=1}^r
      \frac{x_j^2}{(u_\varphi+v_\varphi x_j)^2}
   +\frac1D\sum_{j=1}^r
      \frac{c_\varphi x_j}{(u_\varphi+v_\varphi x_j)^2}
   +o_{\mathbb P}(1)
   \nonumber\\
 =&\frac1D\sum_{j=1}^r
      \frac{x_j^2+c_\varphi x_j}
           {(u_\varphi+v_\varphi x_j)^2}
   +o_{\mathbb P}(1)
   \nonumber\\
 \to&\varphi\left\langle
      \frac{x^2+c_\varphi x}{(u_\varphi+v_\varphi x)^2}
      \right\rangle_{1/\varphi}.
 \label{eq:proof-effective-frobenius-direct}
\end{align}
Here we used \(\mat O_\parallel^\top\mat O_\parallel=\mat I_r\) and the
cyclicity of the trace to combine the two cross terms.  Conditional on
\((x_j,\mat O_\parallel)\), the entries of
\(\mat O_\parallel^\top\mat W\) are independent standard Gaussians.
Consequently, the cross terms in both displays and the final Gaussian term in
the trace-of-the-square display are \(o_{\mathbb P}(1)\).  In the Frobenius
display, \(D^{-1}\mat W^\top\mat W\) instead contributes \(\mat I_r\) inside
the normalized trace.  Equation~\eqref{eq:proof-state-gram-law} then gives the
two displayed limits.

The Marchenko--Pastur equation gives
\begin{equation}
  \left\langle
    \frac1{(1-\alpha)(1-\alpha\varphi)+\alpha x}
  \right\rangle_{1/\varphi}
  =\frac1{1-\alpha\varphi}.
  \label{eq:proof-state-mp-identity}
\end{equation}
Consequently,
\begin{align}
  \left\langle\frac{x}{u_\varphi+v_\varphi x}\right\rangle_{1/\varphi}
  =&\frac1{v_\varphi}
       \left[1-\frac{u_\varphi}{v_\varphi}
          \left\langle\frac1{u_\varphi/v_\varphi+x}\right\rangle_{1/\varphi}\right]\\
  =&\frac1{v_\varphi}[1-(1-\alpha)]\\
  =&\frac{\alpha}{v_\varphi}\\
  =&\frac1{1+(1-\alpha)\varphi}.
\end{align}
The MP equation and its derivative then give
\begin{align}
  \left\langle\frac{x^2}{(u_\varphi+v_\varphi x)^2}\right\rangle_{1/\varphi}
  =&\frac{1+(1-2\alpha)\varphi}
     {[1+(1-\alpha)\varphi]^2(1-\alpha^2\varphi)},\\
  \left\langle\frac{x}{(u_\varphi+v_\varphi x)^2}\right\rangle_{1/\varphi}
  =&\frac1{[1+(1-\alpha)\varphi]^2(1-\alpha^2\varphi)}.
\end{align}
Substitution into Equation~\eqref{eq:proof-effective-trace-direct} gives
\begin{align}
 -\lim_{D\to\infty}
   \frac{\operatorname{tr}\hat{\mat A}_{\mathrm{eff}}}{\eta D}
 =&\frac{\varphi}{1+(1-\alpha)\varphi}.
\end{align}
Substitution into Equation~\eqref{eq:proof-effective-square-direct} gives
\begin{align}
 \lim_{D\to\infty}
   \frac{\operatorname{tr}(\hat{\mat A}_{\mathrm{eff}}^2)}{\eta^2D}
 =&\frac{\varphi[1+(1-2\alpha)\varphi]}
 {[1+(1-\alpha)\varphi]^2(1-\alpha^2\varphi)}.
\end{align}
Finally, substitution into Equation~\eqref{eq:proof-effective-frobenius-direct} gives
\begin{align}
 \lim_{D\to\infty}
   \frac{\|\hat{\mat A}_{\mathrm{eff}}\|_{\mathrm F}^2}{\eta^2D}
 =&\frac{\varphi[1+(1-2\alpha)\varphi+c_\varphi]}
 {[1+(1-\alpha)\varphi]^2(1-\alpha^2\varphi)}\\
 =&\frac{\varphi[1+(1-\alpha)\varphi](1-\alpha^2\varphi)}
 {[1+(1-\alpha)\varphi]^2(1-\alpha^2\varphi)}\\
 =&\frac{\varphi}{1+(1-\alpha)\varphi}.
\end{align}
Hence the three moments agree with Proposition~\ref{prop:cavity-moments}.
\end{proof}

\begin{proposition}[GD recovery for rank-deficient contexts]
\label{prop:rank-deficient-gd-recovery}
Suppose \(0<\alpha<1\), \(\alpha\varphi>1\), and \(\lambda\to0\).
Under the same cavity approximation as in
Proposition~\ref{prop:cavity-moments}, the effective learned matrix satisfies
\begin{equation}
  \hat{\mat A}_{\mathrm{eff}}=-\eta\mat I_D.
  \label{eq:rank-deficient-effective-gd}
\end{equation}
Consequently,
\begin{equation}
  \mathcal E(t)=\mathcal E_{\mathrm{GD}}(t).
\end{equation}
\end{proposition}

\begin{proof}
Let
\begin{equation}
  \mat Z_0:=\lim_{\epsilon\to0}\epsilon\mathcal R_\epsilon,
\end{equation}
where \(\mathcal R_\epsilon\) is defined in
Equation~\eqref{eq:cavity-identification-resolvent}.
Equation~\eqref{eq:proof-cavity-ridge-solution} gives
\begin{equation}
  \operatorname{vec}(\hat{\mat A}_{\mathrm{eff}}+\eta\mat I_D)
  =\eta\mat Z_0\operatorname{vec}\mat I_D.
  \label{eq:proof-identified-effective-matrix}
\end{equation}
Applying the standard Wishart resolvent calculation gives the leading small-\(\epsilon\) equation for \(\rho_\epsilon:=D^{-2}\operatorname{tr}\mathcal R_\epsilon\):
\begin{equation}
  \epsilon\rho_\epsilon^2
  -(1-\alpha\varphi)\rho_\epsilon
  -a_{K,\alpha}=o(1),
  \qquad a_{K,\alpha}>0.
\label{eq:proof-cavity-identification-branches}
\end{equation}
Its positive solution is
\begin{equation}
  \rho_\epsilon
  =\frac{1-\alpha\varphi
    +\sqrt{(1-\alpha\varphi)^2+4\epsilon a_{K,\alpha}}}
    {2\epsilon}+o(1).
\end{equation}
When \(\alpha\varphi>1\), the two terms in the numerator cancel at leading
order and \(\rho_\epsilon\to a_{K,\alpha}/(\alpha\varphi-1)\).  Therefore
\begin{equation}
  \lim_{D\to\infty}\frac1{D^2}\operatorname{tr}\mat Z_0
  =\lim_{\epsilon\to0}\epsilon\rho_\epsilon=0.
\end{equation}
Since \(\mat Z_0\) is the orthogonal projector onto the kernel of
\(\sum_\mu\mat Q_\mu\otimes\mat P_\mu\), this gives
\(\mat Z_0=\mat0\) in the effective large-\(D\) description.
Equation~\eqref{eq:proof-identified-effective-matrix}
then gives Equation~\eqref{eq:rank-deficient-effective-gd}.
Proposition~\ref{prop:gd-recovery-risk} completes the proof.
\end{proof}

For the remainder of the cavity calculation, we therefore use the replacement
\begin{equation}
  \hat{\mat A}\ \longrightarrow\ \hat{\mat A}_{\mathrm{eff}}.
  \label{eq:proof-effective-matrix-replacement}
\end{equation}
To avoid additional notation, we write \(\hat{\mat A}\) for \(\hat{\mat A}_{\mathrm{eff}}\) after Equation~\eqref{eq:proof-effective-matrix-replacement}.

\subsection{Resolvent calculation}

We now use Equation~\eqref{eq:proof-cavity-matrix-profile} to calculate the two resolvents needed to determine \(\varphi_-\) and \(\varphi_+\).
As a preparation, we calculate two scalar functions below:
\begin{align}
  r_\varphi(y)
  :=&\lim_{D\to\infty}\frac1D
      \operatorname{tr}(\mat I_D-y\hat{\mat A}/\eta)^{-1},\\
  G_\varphi(y)
  :=&\lim_{D\to\infty}\frac1D
      \|(\mat I_D-y\hat{\mat A}/\eta)^{-1}\|_{\mathrm F}^2.
\end{align}
These resolvent quantities are given by the following proposition.

\begin{proposition}[Resolvents of the effective matrix]
\label{prop:effective-matrix-resolvents}
For sufficiently small \(y\),
\begin{align}
  r_\varphi(y)
  =&1-\varphi+\varphi
      \left\langle\frac{u_\varphi+v_\varphi x}
       {u_\varphi+(v_\varphi+y)x}\right\rangle_{1/\varphi},
  \label{eq:proof-one-copy-resolvent}\\
  G_\varphi(y)
  =&\frac{1-\varphi+\varphi
      \left\langle (u_\varphi+v_\varphi x)^2/
         [u_\varphi+(v_\varphi+y)x]^2\right\rangle_{1/\varphi}}
       {1-\varphi y^2c_\varphi
      \left\langle x/[u_\varphi+(v_\varphi+y)x]^2
      \right\rangle_{1/\varphi}}.
  \label{eq:proof-two-copy-resolvent}
\end{align}
\end{proposition}

\noindent\emph{Proof.}

\paragraph{Calculation of \(r_\varphi(y)\).}
The effective matrix vanishes on the subspace spanned by \(\mat O_\perp\), so the inverse is the identity on this subspace.
It remains to calculate the inverse on the subspace spanned by \(\mat O_\parallel\).
For sufficiently small \(y\), factoring the deterministic diagonal matrix in Equation~\eqref{eq:proof-cavity-matrix-profile} gives the convergent expansion
\begin{align}
 &\mat O_\parallel^\top
   (\mat I_D-y\hat{\mat A}/\eta)^{-1}\mat O_\parallel\notag\\
 =&\operatorname{diag}_j
   \left(\frac{u_\varphi+v_\varphi x_j}
   {u_\varphi+(v_\varphi+y)x_j}\right)
   \sum_{n=0}^{\infty}\left[
   \frac{y}{\sqrt D}\mat O_\parallel^\top\mat W
   \operatorname{diag}_j\left(
   \frac{\sqrt{c_\varphi x_j}}
   {u_\varphi+(v_\varphi+y)x_j}\right)
   \right]^n.
 \label{eq:proof-one-copy-series}
\end{align}
Conditional on \(\mat O_\parallel\) and \(x_1,\ldots,x_r\), the entries of \(\mat O_\parallel^\top\mat W\) are independent standard Gaussians.
Writing \(i_{n+1}=i_1\), the normalized trace of the \(n\)-th term in Equation~\eqref{eq:proof-one-copy-series} is
\begin{align}
 &\frac{y^n}{D^{1+n/2}}
 \sum_{i_1,\ldots,i_n=1}^r
 \frac{u_\varphi+v_\varphi x_{i_1}}
      {u_\varphi+(v_\varphi+y)x_{i_1}}
 \prod_{k=1}^n
 \frac{\sqrt{c_\varphi x_{i_{k+1}}}}
      {u_\varphi+(v_\varphi+y)x_{i_{k+1}}}
 \prod_{k=1}^n
 (\mat O_\parallel^\top\mat W)_{i_k i_{k+1}}.
\end{align}
For odd \(n\), its conditional Gaussian expectation is zero.
For even \(n\), Wick's rule pairs the factors according to
\begin{equation}
 \E\left[
 (\mat O_\parallel^\top\mat W)_{ij}
 (\mat O_\parallel^\top\mat W)_{kl}
 \right]=\delta_{ik}\delta_{jl}.
\end{equation}
Each pairing leaves at most \(n/2\) free indices in the cyclic product, so for fixed \(n\ge2\) the normalized expected trace is
\begin{equation}
 O\left(\frac{D^{n/2}}{D^{1+n/2}}\right)
 =O(D^{-1})\longrightarrow0.
\end{equation}
Consequently,
\begin{align}
 &\lim_{D\to\infty}\frac1D\E\operatorname{tr}\left\{
   \operatorname{diag}_j\left(
   \frac{u_\varphi+v_\varphi x_j}
   {u_\varphi+(v_\varphi+y)x_j}\right)
   \left[
   \frac{y}{\sqrt D}\mat O_\parallel^\top\mat W
   \operatorname{diag}_j\left(
   \frac{\sqrt{c_\varphi x_j}}
   {u_\varphi+(v_\varphi+y)x_j}\right)
   \right]^n\right\}\notag\\
 =&0,\qquad n\ge1.
 \label{eq:proof-one-copy-higher-terms}
\end{align}
Thus only the \(n=0\) term remains in the normalized trace.
The inactive state directions contribute an identity block, so
\begin{align}
  r_\varphi(y)
  =&(1-\varphi)_+
    +\lim_{D\to\infty}\frac1D\sum_{j=1}^r
       \frac{u_\varphi+v_\varphi x_j}
            {u_\varphi+(v_\varphi+y)x_j}\\
  =&1-\varphi+\varphi
      \left\langle\frac{u_\varphi+v_\varphi x}
       {u_\varphi+(v_\varphi+y)x}\right\rangle_{1/\varphi}.
\end{align}
For \(\varphi>1\), the zero atom contributes \(\varphi-1\), cancelling the negative term \(1-\varphi\).
Thus the same formula also holds for \(\varphi>1\).

\paragraph{Calculation of \(G_\varphi(y)\).}
The squared Frobenius norm contains products of two Gaussian factors, so these terms do not vanish.
We calculate them by the cavity method.
Define
\begin{equation}
  \mat R_y
  :=(\mat I_D-y\hat{\mat A}/\eta)^{-1}.
\end{equation}
Let \(\vb w_j\) and \(\vb o_j\) be the \(j\)-th columns of \(\mat W\) and \(\mat O_\parallel\), respectively.
The cavity step removes \(\vb w_j\) from \(\mat W\) and defines \(\mat R_y^{(j)}\) as the resulting inverse.
Then \(\mat R_y^{(j)}\) is independent of \(\vb w_j\), and
\begin{align}
  \mat R_y^{-1}
  =&(\mat R_y^{(j)})^{-1}
    -\frac{y}{\sqrt D}
      \frac{\sqrt{c_\varphi x_j}}{u_\varphi+v_\varphi x_j}
      \vb w_j\vb o_j^\top.
\end{align}
Restoring the removed column with the Woodbury formula gives
\begin{align}
  \mat R_y
  =&\mat R_y^{(j)}
   +\frac{
      \dfrac{y}{\sqrt D}
      \dfrac{\sqrt{c_\varphi x_j}}{u_\varphi+v_\varphi x_j}
      \mat R_y^{(j)}\vb w_j\vb o_j^\top\mat R_y^{(j)}
    }{
      1-\dfrac{y}{\sqrt D}
      \dfrac{\sqrt{c_\varphi x_j}}{u_\varphi+v_\varphi x_j}
      \vb o_j^\top\mat R_y^{(j)}\vb w_j
    }.
  \label{eq:proof-effective-resolvent-woodbury}
\end{align}
For sufficiently small \(y\), the norm of \(\mat R_y^{(j)}\) is bounded, so \(\vb o_j^\top\mat R_y^{(j)}\vb w_j=O_{\mathbb P}(1)\).
Therefore the denominator is \(1+O_{\mathbb P}(D^{-1/2})\) and converges to one.
Because the cavity matrix is independent of \(\vb w_j\), its conditional average satisfies
\begin{equation}
  \frac1D\E\!\left[
     \vb w_j^\top\mat R_y^{(j)}\mat R_y^{(j)\top}\vb w_j
       \,\middle|\,\mat R_y^{(j)}\right]
  =\frac1D\operatorname{tr}(\mat R_y^{(j)}\mat R_y^{(j)\top})
  \longrightarrow G_\varphi(y).
\end{equation}
The resolvent identity and Equation~\eqref{eq:proof-effective-resolvent-woodbury} give, at the retained leading order,
\begin{align}
 \mat R_y\vb o_j
 =&\frac{u_\varphi+v_\varphi x_j}
         {u_\varphi+(v_\varphi+y)x_j}\vb o_j
   +\frac{y}{\sqrt D}
    \frac{\sqrt{c_\varphi x_j}}
         {u_\varphi+(v_\varphi+y)x_j}\mat R_y\vb w_j\\
 \simeq&\frac{u_\varphi+v_\varphi x_j}
         {u_\varphi+(v_\varphi+y)x_j}\vb o_j
   +\frac{y}{\sqrt D}
    \frac{\sqrt{c_\varphi x_j}}
         {u_\varphi+(v_\varphi+y)x_j}\mat R_y^{(j)}\vb w_j.
\end{align}
The mixed term has zero conditional mean because \(\mat R_y^{(j)}\) is independent of \(\vb w_j\).
Consequently,
\begin{align}
 \E_{\vb w_j}\|\mat R_y\vb o_j\|^2
 \simeq&\frac{(u_\varphi+v_\varphi x_j)^2}
             {[u_\varphi+(v_\varphi+y)x_j]^2}
   +\frac{y^2c_\varphi x_j}
          {D[u_\varphi+(v_\varphi+y)x_j]^2}
     \operatorname{tr}(\mat R_y^{(j)}\mat R_y^{(j)\top}).
\end{align}
Since \(\mat R_y\mat O_\perp=\mat O_\perp\), summing over \(j\) and dividing by \(D\) gives
\begin{align}
 G_\varphi(y)
 =&\lim_{D\to\infty}\left\{
   \frac{D-r}{D}
   +\frac1D\sum_{j=1}^r
     \frac{(u_\varphi+v_\varphi x_j)^2}
          {[u_\varphi+(v_\varphi+y)x_j]^2}
   +\frac{y^2c_\varphi}{D^2}\sum_{j=1}^r
     \frac{x_j\operatorname{tr}(\mat R_y^{(j)}\mat R_y^{(j)\top})}
          {[u_\varphi+(v_\varphi+y)x_j]^2}
 \right\}.
\end{align}
Using \(D^{-1}\operatorname{tr}(\mat R_y^{(j)}\mat R_y^{(j)\top})\to G_\varphi(y)\) and Equation~\eqref{eq:proof-state-gram-law} gives the self-consistent equation
\begin{align}
  G_\varphi(y)
  =&1-\varphi+\varphi
      \left\langle\frac{(u_\varphi+v_\varphi x)^2}
         {[u_\varphi+(v_\varphi+y)x]^2}\right\rangle_{1/\varphi}
   +\varphi y^2c_\varphi
      \left\langle\frac{x}{[u_\varphi+(v_\varphi+y)x]^2}
      \right\rangle_{1/\varphi}G_\varphi(y)\\
  =&
  \frac{1-\varphi+\varphi
      \left\langle (u_\varphi+v_\varphi x)^2/
         [u_\varphi+(v_\varphi+y)x]^2\right\rangle_{1/\varphi}}
       {1-\varphi y^2c_\varphi
      \left\langle x/[u_\varphi+(v_\varphi+y)x]^2
      \right\rangle_{1/\varphi}}.
\end{align}
The first equality is the self-consistent equation for \(G_\varphi(y)\), and the second equality solves it.
This proves the second formula in Proposition~\ref{prop:effective-matrix-resolvents}.
\hfill\(\square\)\par

\subsection{The spectrum conditions}
\label{subsec:spectrum-conditions}
In this subsection, we first express the long-time behavior in terms of the spectral radius of the inference matrix.
Define \(\vb e_t:=\hat{\vb w}_t-\vb w^{\mathrm{in}}\).
For a fixed test context, Equation~\eqref{eq:learned-iteration} gives the inference recursion
\begin{equation}
  \begin{aligned}
  \vb e_{t+1}
  =&\mat J\vb e_t
   +\mat H^{\mathrm{in}}(\hat{\mat A}+\eta\mat I_D)
      \vb w^{\mathrm{in}}
   +\frac{\eta}{L}\mat X^{\mathrm{in}}\vb\varepsilon^{\mathrm{in}},
  \end{aligned}
\end{equation}
where
\begin{equation}
  \mat J:=\mat I_D+\mat H^{\mathrm{in}}\hat{\mat A}.
\end{equation}
Therefore,
\begin{equation}
  \begin{aligned}
  \vb e_t
  =&\mat J^t\vb e_0
   +\sum_{s=0}^{t-1}\mat J^s
      \left[
        \mat H^{\mathrm{in}}(\hat{\mat A}+\eta\mat I_D)
           \vb w^{\mathrm{in}}
        +\frac{\eta}{L}\mat X^{\mathrm{in}}
           \vb\varepsilon^{\mathrm{in}}
      \right].
  \end{aligned}
\end{equation}
The long-time behavior is consequently determined by
\begin{equation}
  \rho(\mat J)
  :=\max_{\lambda\in\operatorname{spec}(\mat J)}|\lambda|.
\end{equation}
More precisely,
\begin{equation}
  \begin{cases}
    \rho(\mat J)<1
    &\Longrightarrow\quad \mat J^t\to\mat0
       \text{ and the error remains bounded},\\[2mm]
    \rho(\mat J)>1
    &\Longrightarrow\quad \mat J^t
       \text{ and the error grow exponentially},\\[2mm]
    \rho(\mat J)=1
    &\Longrightarrow\quad \text{the limiting case}.
  \end{cases}
\end{equation}
When \(\rho(\mat J)=1\), this gives
\begin{equation}
  \sum_{s=0}^{t-1}\mat J^s
  \left[
    \mat H^{\mathrm{in}}(\hat{\mat A}+\eta\mat I_D)
       \vb w^{\mathrm{in}}
    +\frac{\eta}{L}\mat X^{\mathrm{in}}
       \vb\varepsilon^{\mathrm{in}}
  \right]
  =O(t),
\end{equation}
so the state grows linearly and its squared error grows as \(\Theta(t^2)\).

We next rewrite the conditions under which an eigenvalue of \(\mat J\) reaches \(1\) as conditions on \(\mat A_c\).
Define \(\mat U\) and \(\mat A_c\) by
\begin{equation}
  \mat U:=\mat X^{\mathrm{in}}
      [(\mat X^{\mathrm{in}})^\top\mat X^{\mathrm{in}}]^{-1/2},
  \qquad
  \mat A_c:=\mat U^\top\hat{\mat A}\mat U.
\end{equation}
\begin{figure}[t]
  \centering
  \includegraphics[width=\linewidth]{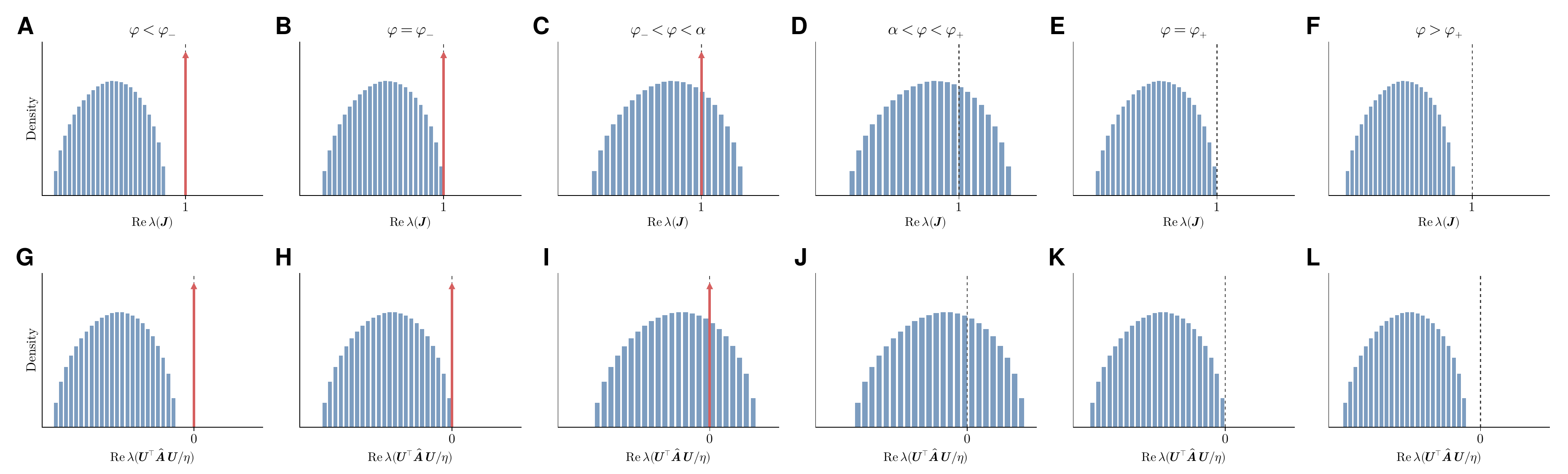}
  \caption{
    \textbf{Schematic spectral evolution across the long-time phases.}
    (A--F) Densities of the real parts of the eigenvalues of
    \(\mat I_D+\mat H^{\mathrm{in}}\hat{\mat A}\).
    (G--L) Densities of the real parts of the eigenvalues of
    \(\mat A_c/\eta\).
    Red arrows in (A--C) and (G--I) mark the rank-deficient atoms at
    eigenvalues \(1\) and \(0\), respectively.
  }
  \label{fig:eigenvalue-phase-schematic}
\end{figure}

The top row of Figure~\ref{fig:eigenvalue-phase-schematic} shows the corresponding changes of the spectrum of \(\mat J\).
In Figure~\ref{fig:eigenvalue-phase-schematic}A and B, \(\rho(\mat J)=1\).
In Figure~\ref{fig:eigenvalue-phase-schematic}C and D, an eigenvalue lies beyond \(1\), and hence \(\rho(\mat J)>1\).
In Figure~\ref{fig:eigenvalue-phase-schematic}E, the edge returns to \(1\), while Figure~\ref{fig:eigenvalue-phase-schematic}F has \(\rho(\mat J)<1\).
The bottom row of Figure~\ref{fig:eigenvalue-phase-schematic} shows the eigenvalues of \(\mat A_c/\eta\).
Since \(\mat U^\top\mat U=\mat I_L\) and \(\mat H^{\mathrm{in}}=\mat X^{\mathrm{in}}(\mat X^{\mathrm{in}})^\top/L\), we have
\begin{align}
  \mat H^{\mathrm{in}}
  =&\mat U
    \frac{(\mat X^{\mathrm{in}})^\top\mat X^{\mathrm{in}}}{L}
    \mat U^\top,\\
  \operatorname{spec}(\mat H^{\mathrm{in}}\hat{\mat A})
       \setminus\{0\}
  =&\operatorname{spec}\left(
      \frac{(\mat X^{\mathrm{in}})^\top\mat X^{\mathrm{in}}}{L}
      \mat A_c
    \right)\setminus\{0\}.
\end{align}
The matrix \((\mat X^{\mathrm{in}})^\top\mat X^{\mathrm{in}}/L\) is positive definite for \(0<\alpha<1\).
Consequently,
\begin{align}
 &\text{an additional eigenvalue of \(\mat A_c\) reaches \(0\)}
 \notag\\
 \Longleftrightarrow\;&
 \text{an additional eigenvalue of
 \(\mat H^{\mathrm{in}}\hat{\mat A}\) reaches \(0\)}
 \notag\\
 \Longleftrightarrow\;&
 \text{an additional eigenvalue of
 \(\mat I_D+\mat H^{\mathrm{in}}\hat{\mat A}\) reaches \(1\)}.
 \label{eq:proof-boundary-eigenvalue-equivalence}
\end{align}

Let
\begin{equation}
  r_D:=\operatorname{rank}(\hat{\mat A})
  =\min\{D,M(K-1)\}.
\end{equation}
If the singular values are ordered as
\begin{equation}
  \sigma_1(\mat A_c)\ge\cdots\ge\sigma_L(\mat A_c)\ge0,
\end{equation}
we write
\begin{equation}
  \sigma_{\min}^{\mathrm{nonzero}}(\mat A_c)
  :=\sigma_{\min\{L,r_D\}}(\mat A_c).
\end{equation}
For \(0<\varphi<\alpha\), the red arrows in Figure~\ref{fig:eigenvalue-phase-schematic}A--C and G--I show the \((\alpha-\varphi)D+o(D)\) eigenvalues fixed at \(1\) and \(0\), respectively.
These fixed eigenvalues do not define the lower boundary.
After excluding them, the lower-boundary condition is
\begin{equation}
  \lim_{D\to\infty}
  \sigma_{\min}^{\mathrm{nonzero}}(\mat A_c)=0.
\end{equation}
This is shown in Figure~\ref{fig:eigenvalue-phase-schematic}B and H, where the edge of the remaining eigenvalues reaches \(1\) in the top row and \(0\) in the bottom row.
For \(\alpha<\varphi<1/\alpha\), there are no fixed zero eigenvalues.
The upper-boundary condition is therefore
\begin{equation}
  \lim_{D\to\infty}
  \sigma_L(\mat A_c)=0.
\end{equation}
This is shown in Figure~\ref{fig:eigenvalue-phase-schematic}E and K, where the edge reaches \(1\) in the top row and \(0\) in the bottom row.

\subsection{Lower boundary}
In this subsection, we determine the lower boundary \(\varphi_-(\alpha)\).
In this regime, \(\mat A_c\) already has \((\alpha-\varphi)D+o(D)\) zero eigenvalues because of its rank deficit.
Therefore, \(\varphi_-(\alpha)\) is the value at which an additional eigenvalue reaches zero.

Equation~\eqref{eq:proof-cavity-matrix-profile} gives the effective matrix, and Equations~\eqref{eq:proof-one-copy-resolvent} and \eqref{eq:proof-two-copy-resolvent} give the two resolvents.
Next, we determine the lower boundary while retaining the Gaussian test covariance.

\begin{proposition}[Lower boundary]
\label{prop:lower-boundary}
Suppose \(0<\alpha<1\).
There is a unique \(\varphi_-(\alpha)\in(0,\alpha)\) such that
\begin{equation}
  \lim_{D\to\infty}
  \sigma_{\min}^{\mathrm{nonzero}}(\mat A_c)=0
  \quad\Longleftrightarrow\quad
  \varphi=\varphi_-(\alpha),
  \qquad 0<\varphi<\alpha.
\end{equation}
It satisfies
\begin{equation}
  (\alpha-\varphi_-)(1-\varphi_-)
  =\alpha(1-\alpha)\varphi_-^2(1-\alpha\varphi_-)
       (\alpha+1-2\varphi_-),
  \label{eq:phase-three-lower-boundary}
\end{equation}
\end{proposition}

\begin{proof}
We derive Equation~\eqref{eq:phase-three-lower-boundary} for \(0<\varphi<\alpha\) and then show that it has a unique solution.
The matrix \(\mat H^{\mathrm{in}}\hat{\mat A}\) has \((\alpha-\varphi)D+o(D)\) zero eigenvalues in this regime, so replacing \(\mat H^{\mathrm{in}}\) by an orthogonal projector would change its remaining eigenvalues.
In the orthogonal basis \((\mat O_\parallel,\mat O_\perp)\) defined above, define \(\mat X_1\) and \(\mat X_2\) by
\begin{equation}
  \begin{pmatrix}
    \mat O_\parallel^\top\\
    \mat O_\perp^\top
  \end{pmatrix}
  \mat X^{\mathrm{in}}
  =\begin{pmatrix}\mat X_1\\\mat X_2\end{pmatrix}.
\end{equation}
Equation~\eqref{eq:proof-cavity-matrix-profile} gives
\begin{equation}
  \frac1\eta
  \begin{pmatrix}
    \mat O_\parallel^\top\\
    \mat O_\perp^\top
  \end{pmatrix}
  \hat{\mat A}
  \begin{pmatrix}\mat O_\parallel&\mat O_\perp\end{pmatrix}
  =\begin{pmatrix}\mat C&\mat0\\\mat Z&\mat0\end{pmatrix}.
\end{equation}
Here \(\mat X_1\in\mathbb R^{r\times L}\), \(\mat X_2\in\mathbb R^{(D-r)\times L}\), \(\mat C\in\mathbb R^{r\times r}\), and \(\mat Z\in\mathbb R^{(D-r)\times r}\), where \(r/D\to\varphi<\alpha\).
The matrix \(\mat C\) is invertible almost surely, so define
\begin{equation}
  \mat R:=\mat Z\mat C^{-1}.
\end{equation}
Since \(r<L\), the matrix
\begin{equation}
  \mat B:=\frac{\mat X_1\mat X_1^\top}{L}
\end{equation}
is also invertible almost surely.
Direct multiplication gives
\begin{align}
 \frac1\eta
  \begin{pmatrix}
    \mat O_\parallel^\top\\
    \mat O_\perp^\top
  \end{pmatrix}
  \mat H^{\mathrm{in}}\hat{\mat A}
  \begin{pmatrix}\mat O_\parallel&\mat O_\perp\end{pmatrix}
 =&\frac1L
  \begin{pmatrix}
    \mat X_1\mat X_1^\top&\mat X_1\mat X_2^\top\\
    \mat X_2\mat X_1^\top&\mat X_2\mat X_2^\top
  \end{pmatrix}
  \begin{pmatrix}\mat C&\mat0\\\mat Z&\mat0\end{pmatrix}\\
 =&\frac1L
  \begin{pmatrix}
    \mat X_1\mat X_1^\top\mat C
      +\mat X_1\mat X_2^\top\mat Z&\mat0\\
    \mat X_2\mat X_1^\top\mat C
      +\mat X_2\mat X_2^\top\mat Z&\mat0
  \end{pmatrix}.
\end{align}
Therefore,
\begin{align}
 \det\left(\lambda\mat I_D-\frac{\mat H^{\mathrm{in}}\hat{\mat A}}\eta\right)
 =&\lambda^{D-r}
   \det\left[
     \lambda\mat I_r
     -\frac1L\left(
       \mat X_1\mat X_1^\top\mat C
       +\mat X_1\mat X_2^\top\mat Z
     \right)
   \right].
\end{align}
Hence the nonzero eigenvalues of
\(\mat H^{\mathrm{in}}\hat{\mat A}/\eta\) are those of the following
\(r\times r\) matrix:
\begin{align}
  \frac1{L}
     (\mat X_1\mat X_1^\top\mat C+\mat X_1\mat X_2^\top\mat Z)
  =&\mat B\mat C+
      \frac{\mat X_1\mat X_2^\top}{L}\mat Z\\
  =&\mat B\mat C+
      \frac{\mat X_1\mat X_2^\top}{L}\mat R\mat C\\
  =&\left(
      \mat B+
      \frac{\mat X_1\mat X_2^\top}{L}\mat R
    \right)\mat C\\
  =&\mat B\left(
      \mat I_r+\mat B^{-1}
      \frac{\mat X_1\mat X_2^\top}{L}\mat R
    \right)\mat C.
  \label{eq:proof-gaussian-nonzero-eigenvalues}
\end{align}
Equation~\eqref{eq:proof-cavity-matrix-profile} gives
\begin{align}
  \mat R
  =&\mat Z\mat C^{-1}\\
  =&\frac{\mat O_\perp^\top\mat W}{\sqrt D}
    \operatorname{diag}_j
    \frac{\sqrt{c_\varphi x_j}}{u_\varphi+v_\varphi x_j}
    \left[
      -\operatorname{diag}_j
       \frac{x_j}{u_\varphi+v_\varphi x_j}
      +\frac{\mat O_\parallel^\top\mat W}{\sqrt D}
       \operatorname{diag}_j
       \frac{\sqrt{c_\varphi x_j}}
            {u_\varphi+v_\varphi x_j}
    \right]^{-1}\\
  =&-\sqrt{\frac{c_\varphi}{D}}\,
    \mat O_\perp^\top\mat W
    \left[
      \operatorname{diag}_j\sqrt{x_j}
      -\sqrt{\frac{c_\varphi}{D}}\,
       \mat O_\parallel^\top\mat W
    \right]^{-1}\\
  =&-\sqrt{\frac{c_\varphi}{D}}\,
    \mat O_\perp^\top\mat W
    \sum_{k=0}^{\infty}
    \left[
      \sqrt{\frac{c_\varphi}{D}}\,
      \operatorname{diag}_j(x_j^{-1/2})
      \mat O_\parallel^\top\mat W
    \right]^k
    \operatorname{diag}_j(x_j^{-1/2}).
\end{align}
The average of terms containing an odd total number of centered Gaussian factors is zero.
At the retained leading order, each additional Gaussian pair multiplies the preceding contribution by \(\varphi c_\varphi\langle x^{-1}\rangle_{1/\varphi}\).
Therefore,
\begin{align}
  \lim_{D\to\infty}\frac{\|\mat R\|_{\mathrm F}^2}{D}
  =&(1-\varphi)\varphi c_\varphi
    \langle x^{-1}\rangle_{1/\varphi}
    \left[
      1
      +\varphi c_\varphi
       \langle x^{-1}\rangle_{1/\varphi}
      +\left(
        \varphi c_\varphi
        \langle x^{-1}\rangle_{1/\varphi}
       \right)^2
      +\cdots
    \right]\\
  =&(1-\varphi)\varphi c_\varphi
    \langle x^{-1}\rangle_{1/\varphi}
    \sum_{j=0}^{\infty}
    \left[
      \varphi c_\varphi
      \langle x^{-1}\rangle_{1/\varphi}
    \right]^j\\
  =&(1-\varphi)\varphi c_\varphi
    \frac{1}{1-\varphi}
    \sum_{j=0}^{\infty}
    \left(
      \frac{\varphi c_\varphi}{1-\varphi}
    \right)^j\\
  =&\varphi c_\varphi
    \sum_{j=0}^{\infty}
    \left(
      \frac{\varphi c_\varphi}{1-\varphi}
    \right)^j\\
  =&\frac{\varphi c_\varphi(1-\varphi)}
           {1-\varphi-\varphi c_\varphi}.
  \label{eq:proof-lower-inverse-moment}
\end{align}
The denominator is positive for \(0<\varphi\le\alpha\).
Indeed, \(\varphi+\varphi c_\varphi\) is increasing there and
\begin{equation}
  1-\alpha-\alpha c_\alpha
  =(1-\alpha)[1-\alpha^3(1-\alpha^2)]>0.
\end{equation}

Conditional on \(\mat X_1\) and the learned matrix, \(\mat B^{-1}\mat X_1\mat X_2^\top\mat R/L\) is centered Gaussian.
For any matrix \(\mat F\) independent of \(\mat X_2\), direct averaging over \(\mat X_2\) gives
\begin{align}
  \E_{\mat X_2}\left[
    \mat B^{-1}\frac{\mat X_1\mat X_2^\top}{L}
    \mat R\mat F\mat R^\top
    \frac{\mat X_2\mat X_1^\top}{L}\mat B^{-1}
  \right]
  =&\frac{\mat B^{-1}}{L}
       \operatorname{tr}(\mat R^\top\mat R\mat F).
\end{align}
The right-hand side is proportional to \(\mat B^{-1}\).
Substituting \(\mat F=\mat B^{-1}\) into the same equality gives
\begin{align}
 \frac{\mat B^{-1}}L
 \operatorname{tr}(\mat R^\top\mat R\mat B^{-1})
 =&\left\{\frac1L
 \operatorname{tr}(\mat B^{-1}\mat R^\top\mat R)\right\}\mat B^{-1}.
\end{align}
Thus each additional pair of centered Gaussian factors multiplies the preceding term by \(L^{-1}\operatorname{tr}(\mat B^{-1}\mat R^\top\mat R)\).
Rotational independence and the Wishart inverse moment give
\begin{align}
  \frac1{L}
      \operatorname{tr}(\mat B^{-1}\mat R^\top\mat R)
  \longrightarrow&
      \frac D{L}
      \left(\frac1r\operatorname{tr}\mat B^{-1}\right)
      \left(\frac1D\operatorname{tr}\mat R^\top\mat R\right)\\
  =&\frac1\alpha\frac{\alpha}{\alpha-\varphi}
      \frac{\varphi c_\varphi(1-\varphi)}
           {1-\varphi-\varphi c_\varphi}\\
  =&\frac{\varphi c_\varphi(1-\varphi)}
       {(\alpha-\varphi)(1-\varphi-\varphi c_\varphi)}.
  \label{eq:proof-lower-pair-ratio}
\end{align}
Since \(\mat B\) and \(\mat C\) are invertible almost surely.
Its inverse has the expansion
\begin{align}
  \left(
    \mat I_r+\mat B^{-1}
    \frac{\mat X_1\mat X_2^\top}{L}\mat R
  \right)^{-1}
  =&\mat I_r
  -\mat B^{-1}
    \frac{\mat X_1\mat X_2^\top}{L}\mat R
  +\left(
    \mat B^{-1}
    \frac{\mat X_1\mat X_2^\top}{L}\mat R
  \right)^2
  -\cdots.
\end{align}
At the retained leading order, the Frobenius norm of this expansion is a geometric series whose ratio is given by Equation~\eqref{eq:proof-lower-pair-ratio}.
In the effective random-matrix calculation, the lower boundary is therefore identified by the divergence of this inverse norm, namely
\begin{align}
  1
  =&\frac1L
      \operatorname{tr}(\mat B^{-1}\mat R^\top\mat R)\\
  =&\frac{\varphi c_\varphi(1-\varphi)}
       {(\alpha-\varphi)(1-\varphi-\varphi c_\varphi)},
\end{align}
which gives
\begin{align}
  (\alpha-\varphi)(1-\varphi)
  =\alpha(1-\alpha)\varphi^2(1-\alpha\varphi)(\alpha+1-2\varphi).
  \label{eq:self-con-phi-p}
\end{align}
Thus the boundary equation is Equation~\eqref{eq:phase-three-lower-boundary}.

It remains to show that Equation~\eqref{eq:self-con-phi-p} has a unique solution in ($0<\varphi<\alpha$).
To check uniqueness directly, set \(\varphi=\alpha z/(1+z)\).
Multiplying the left side minus the right side of Equation~\eqref{eq:phase-three-lower-boundary} by \((1+z)^4\) gives
\begin{align}
  0=&\alpha+\alpha(3-\alpha)z
      +\alpha[(1-\alpha)(3+\alpha)+\alpha^4]z^2\notag\\
     &+\alpha(1-\alpha)[(1-\alpha^2)^2+\alpha^5]z^3
      -\alpha^3(1-\alpha)^3(1+\alpha)z^4.
\end{align}
There is exactly one sign change, so Descartes' rule gives one positive root.
Since \(z>0\) parametrizes \(0<\varphi<\alpha\), the lower boundary is unique in that interval.
\end{proof}

\subsection{Upper boundary}
In this subsection, we determine the upper boundary \(\varphi_+(\alpha)\).
In this regime, \(\mat A_c\) is generically full rank.
Therefore, \(\varphi_+(\alpha)\) is the value at which its smallest singular
value reaches zero, which in our effective random-matrix description is
expected to coincide with the limiting eigenvalue support reaching zero.

The preceding calculation gives the value at which another zero eigenvalue appears for \(0<\varphi<\alpha\).
We next calculate the corresponding value for \(\alpha<\varphi<1/\alpha\) by reducing the two resolvent conditions to one scalar MP equation.

\begin{proposition}[Upper boundary]
\label{prop:upper-boundary}
Suppose \(0<\alpha<1\).
Let \(z_+\) be the unique positive solution of
\begin{align}
  0=&(1-\alpha)^2z_+^4
    +(1-\alpha)[3(1-\alpha^2)+\alpha^3]z_+^3\notag\\
    &+(1-\alpha)[3+3\alpha(1-\alpha)+\alpha^4]z_+^2\notag\\
    &+[(1-\alpha)(1+2\alpha)+\alpha^5]z_+
      -\alpha^4(1-\alpha^2).
  \label{eq:phase-three-upper-polynomial}
\end{align}
Then
\begin{equation}
  \varphi_+(\alpha)
  :=\frac{\alpha^2+z_++(1-\alpha)z_+^2}
          {\alpha(1+z_+)[1+(1-\alpha)z_+]}.
  \label{eq:phase-three-upper-boundary}
\end{equation}
is the unique value in \((\alpha,1/\alpha)\) such that
\begin{equation}
  \lim_{D\to\infty}
  \sigma_L(\mat A_c)=0
  \quad\Longleftrightarrow\quad
  \varphi=\varphi_+(\alpha),
  \qquad \alpha<\varphi<1/\alpha.
\end{equation}
Moreover, \(\alpha<\varphi_+(\alpha)<1\).
\end{proposition}

\begin{proof}
For \(\alpha<\varphi<1/\alpha\), we use that \(\mat A_c\) has full rank except at the boundary calculated below.
The positive-definite matrix \((\mat X^{\mathrm{in}})^\top\mat X^{\mathrm{in}}/L\) is boundedly invertible.
Therefore \(((\mat X^{\mathrm{in}})^\top\mat X^{\mathrm{in}}/L)\mat A_c\) has a zero eigenvalue exactly when \(\mat A_c\) has a zero eigenvalue.
We now obtain the two scalar boundary equations by a leave-one-column cavity calculation.
Let \(\mat U^{-\ell}\) be \(\mat U\) with column \(\vb u_\ell\) removed, and define
\begin{align}
 \mathcal G_\zeta
 :=&\left[\zeta\mat I_L
   -\mat U^\top\frac{\hat{\mat A}}\eta\mat U\right]^{-1},\\
 \mathcal G_\zeta^{-\ell}
 :=&\left[\zeta\mat I_{L-1}
   -(\mat U^{-\ell})^\top
     \frac{\hat{\mat A}}\eta\mat U^{-\ell}\right]^{-1}.
\end{align}
Ordering \(\vb u_\ell\) last gives
\begin{align}
 \mathcal G_\zeta^{-1}
 =&\begin{pmatrix}
   (\mathcal G_\zeta^{-\ell})^{-1}
   &-(\mat U^{-\ell})^\top(\hat{\mat A}/\eta)\vb u_\ell\\
   -\vb u_\ell^\top(\hat{\mat A}/\eta)\mat U^{-\ell}
   &\zeta-\vb u_\ell^\top(\hat{\mat A}/\eta)\vb u_\ell
 \end{pmatrix}.
\end{align}
Restoring the deleted column with the Woodbury formula gives
\begin{align}
 [\mathcal G_\zeta]_{\ell\ell}
 =&\left[\zeta
   -\vb u_\ell^\top\frac{\hat{\mat A}}\eta\vb u_\ell
   -\vb u_\ell^\top\frac{\hat{\mat A}}\eta\mat U^{-\ell}
    \mathcal G_\zeta^{-\ell}(\mat U^{-\ell})^\top
    \frac{\hat{\mat A}}\eta\vb u_\ell
 \right]^{-1}.
 \label{eq:proof-upper-column-woodbury}
\end{align}
We use the large-\(D\) replacement that \(\vb u_\ell\) can be averaged independently of \(\mathcal G_\zeta^{-\ell}\).
Define the scalar order parameter by
\begin{equation}
 y:=\lim_{D\to\infty}\lim_{\zeta\to0}
    \frac1D\operatorname{tr}\mathcal G_\zeta.
\end{equation}
Averaging the left-hand side of Equation~\eqref{eq:proof-upper-column-woodbury} over \(\ell\) gives
\begin{align}
 \lim_{D\to\infty}\lim_{\zeta\to0}
 \frac1L\sum_{\ell=1}^L[\mathcal G_\zeta]_{\ell\ell}
 =&\lim_{D\to\infty}\lim_{\zeta\to0}
 \frac1L\operatorname{tr}\mathcal G_\zeta\\
 =&\lim_{D\to\infty}\frac DL
 \left(\lim_{\zeta\to0}\frac1D
 \operatorname{tr}\mathcal G_\zeta\right)\\
 =&\frac y\alpha.
\end{align}
Applying cavity independence and concentration on the right-hand side of Equation~\eqref{eq:proof-upper-column-woodbury}, and then averaging over \(\ell\), gives
\begin{align}
 &\lim_{D\to\infty}\lim_{\zeta\to0}
 \frac1L\sum_{\ell=1}^L
 \left[
 \zeta
   -\vb u_\ell^\top\frac{\hat{\mat A}}\eta\vb u_\ell
   -\vb u_\ell^\top\frac{\hat{\mat A}}\eta\mat U^{-\ell}
    \mathcal G_\zeta^{-\ell}(\mat U^{-\ell})^\top
    \frac{\hat{\mat A}}\eta\vb u_\ell
 \right]^{-1}\\
 \simeq&\left\{
 -\lim_{D\to\infty}\lim_{\zeta\to0}
 \frac1{LD}\sum_{\ell=1}^L
 \operatorname{tr}\left[
   \frac{\hat{\mat A}}\eta
   +\frac{\hat{\mat A}}\eta\mat U^{-\ell}
    \mathcal G_\zeta^{-\ell}(\mat U^{-\ell})^\top
    \frac{\hat{\mat A}}\eta
 \right]
 \right\}^{-1}\\
 \simeq&\left\{
 -\lim_{D\to\infty}\frac1D
 \operatorname{tr}\left[
   \frac{\hat{\mat A}}\eta
   +y\frac{\hat{\mat A}}\eta
    (\mat I_D-y\hat{\mat A}/\eta)^{-1}
    \frac{\hat{\mat A}}\eta
 \right]
 \right\}^{-1}\\
 =&\left\{
 -\lim_{D\to\infty}\frac1D
 \operatorname{tr}\left[
   \frac{\hat{\mat A}}\eta
   +y\left(\frac{\hat{\mat A}}\eta\right)^2
    \sum_{k=0}^{\infty}
    \left(y\frac{\hat{\mat A}}\eta\right)^k
 \right]
 \right\}^{-1}\\
 =&\left\{
 -\lim_{D\to\infty}\frac1D
 \operatorname{tr}\left[
   \frac{\hat{\mat A}}\eta
   +y\left(\frac{\hat{\mat A}}\eta\right)^2
   +y^2\left(\frac{\hat{\mat A}}\eta\right)^3
   +\cdots\right]
 \right\}^{-1}.
\end{align}
Comparing the averages of the left- and right-hand sides of Equation~\eqref{eq:proof-upper-column-woodbury} and taking their inverses gives
\begin{align}
 \frac\alpha y
 \simeq&-\lim_{D\to\infty}\frac1D
 \operatorname{tr}\left[
   \frac{\hat{\mat A}}\eta
   \sum_{k=0}^{\infty}
   \left(y\frac{\hat{\mat A}}\eta\right)^k
 \right]
 =-\lim_{D\to\infty}\frac1D\operatorname{tr}\left[
   \frac{\hat{\mat A}}\eta
   (\mat I_D-y\hat{\mat A}/\eta)^{-1}\right].
\end{align}
Multiplying by \(y\) and using Proposition~\ref{prop:effective-matrix-resolvents} gives
\begin{align}
 \alpha
 =&-\lim_{D\to\infty}\frac1D\operatorname{tr}\left[
   y\frac{\hat{\mat A}}\eta
   (\mat I_D-y\hat{\mat A}/\eta)^{-1}\right]\\
 =&-\lim_{D\to\infty}\frac1D\operatorname{tr}\left[
   (\mat I_D-y\hat{\mat A}/\eta)^{-1}-\mat I_D\right]\\
 =&1-r_\varphi(y).
\end{align}
Hence
\begin{equation}
 r_\varphi(y)=1-\alpha.
 \label{eq:proof-upper-first-boundary-equation}
\end{equation}
To obtain the second boundary condition, vary \(\mathcal G_\zeta^{-\ell}\) in Equation~\eqref{eq:proof-upper-column-woodbury} while keeping \(\vb u_\ell\) fixed.
The linear variation satisfies
\begin{align}
 \delta[\mathcal G_\zeta]_{\ell\ell}
 =&[\mathcal G_\zeta]_{\ell\ell}^{2}
   \vb u_\ell^\top\frac{\hat{\mat A}}\eta\mat U^{-\ell}
   (\delta\mathcal G_\zeta^{-\ell})(\mat U^{-\ell})^\top
   \frac{\hat{\mat A}}\eta\vb u_\ell.
\end{align}
At the upper boundary, the nonzero eigenvalue support of \(\mat A_c/\eta\) reaches zero, as shown in Figure~\ref{fig:eigenvalue-phase-schematic}K; the corresponding edge of the inference matrix reaches one in Figure~\ref{fig:eigenvalue-phase-schematic}E.
The resolvent \(\mathcal G_\zeta\) then becomes sensitive to a small change in the cavity equation.
Using the same scalar \(y\) as in the first condition, this criterion becomes
\begin{align}
 1
 \simeq&\lim_{D\to\infty}\frac1L\left\|
   y\frac{\hat{\mat A}}\eta
   (\mat I_D-y\hat{\mat A}/\eta)^{-1}\right\|_{\mathrm F}^2\\
 =&\frac1\alpha\lim_{D\to\infty}\frac1D\left\|
   (\mat I_D-y\hat{\mat A}/\eta)^{-1}-\mat I_D
   \right\|_{\mathrm F}^2\\
 =&\frac{G_\varphi(y)-2r_\varphi(y)+1}{\alpha}.
\end{align}
Substituting Equation~\eqref{eq:proof-upper-first-boundary-equation} gives
\begin{equation}
 G_\varphi(y)=1-\alpha.
 \label{eq:proof-upper-second-boundary-equation}
\end{equation}
Thus the upper boundary is determined by
\begin{equation}
 r_\varphi(y)=1-\alpha,
 \qquad
 G_\varphi(y)=1-\alpha,
 \qquad y>0.
 \label{eq:proof-upper-two-boundary-equations}
\end{equation}

For the scalar elimination, put
\begin{equation}
  \beta:=v_\varphi+y,
  \qquad
  m:=\varphi\left\langle\frac{\beta x}{u_\varphi+\beta x}
                  \right\rangle_{1/\varphi}.
\end{equation}
The first resolvent equation reduces to
\begin{align}
  1-\alpha
  =&1-\varphi+\varphi
       \left\langle1-\frac{yx}{u_\varphi+\beta x}\right\rangle_{1/\varphi}\\
  =&1-\frac y\beta\,m,
\end{align}
hence $\frac y\beta=\frac{\alpha}{m}$.
The MP identity is
\begin{equation}
  \frac{u_\varphi}{\beta}m=(\varphi-m)(1-m).
\end{equation}
Using \(y/\beta=\alpha/m\), it becomes
\begin{align}
  (\varphi-m)(1-m)
  =\frac{u_\varphi}{v_\varphi}(m-\alpha)
  =\frac{(1-\alpha)(1-\alpha\varphi)}{\alpha}(m-\alpha)
  \label{eq:proof-upper-mp-equation}
\end{align}
for $\alpha<m<\min\{\varphi,1\}$.

We next reduce the Frobenius-resolvent equation
\eqref{eq:proof-two-copy-resolvent} explicitly.
Differentiating the MP identity at fixed \(u_\varphi\) gives
\begin{align}
  \beta\partial_\beta m
  =&\frac{(u_\varphi/\beta)m}
        {u_\varphi/\beta+\varphi+1-2m}
  =\frac{m(\varphi-m)(1-m)}{\varphi-m^2},\\
  \varphi\left\langle
     \frac{(\beta x)^2}{(u_\varphi+\beta x)^2}
     \right\rangle_{1/\varphi}
  =&m-\beta\partial_\beta m.
\end{align}
Subtracting \(1-\alpha\) from Equation~\eqref{eq:proof-two-copy-resolvent} and multiplying by its denominator yields
\begin{align}
  &\left[1-\varphi y^2c_\varphi
       \left\langle\frac{x}{(u_\varphi+\beta x)^2}\right\rangle_{1/\varphi}\right]
       [G_\varphi(y)-(1-\alpha)]\notag\\
  \quad=&-\alpha+\frac{\alpha^2}{m}
       -\frac{\alpha^2}{m^2}\beta\partial_\beta m
       +\frac{(1-\alpha)\alpha^2c_\varphi}
              {m^2(u_\varphi/\beta)}\beta\partial_\beta m\\
  \quad=&\frac{\alpha}{\varphi-m^2}
       [m^2-2\alpha m+\alpha-(1-\alpha)\varphi
           +\alpha(1-\alpha)c_\varphi]\\
  \quad=&-\frac{\alpha(1-\alpha)}{\varphi-m^2}
       \left[\varphi-\alpha-\alpha c_\varphi
                 -\frac{(m-\alpha)^2}{1-\alpha}\right].
\end{align}
Thus the upper boundary satisfies the scalar equation
\begin{equation}
  \varphi-\alpha
  =\alpha c_\varphi+\frac{(m-\alpha)^2}{1-\alpha}.
  \label{eq:proof-upper-scalar-boundary}
\end{equation}

Set \(z=(m-\alpha)/(1-m)>0\), so that \(m=(\alpha+z)/(1+z)\).
Solving Equation~\eqref{eq:proof-upper-mp-equation} for \(\varphi\) gives
\begin{align}
  \varphi
  =&\frac{\alpha m(1-m)+(1-\alpha)(m-\alpha)}
       {\alpha[(1-m)+(1-\alpha)(m-\alpha)]}\\
  =&\frac{\alpha^2+z+(1-\alpha)z^2}
       {\alpha(1+z)[1+(1-\alpha)z]}.
  \label{eq:proof-upper-parametrization}
\end{align}
Substitution into Equation~\eqref{eq:proof-upper-scalar-boundary} gives
\begin{align*}
  0=&(1-\alpha)^2z^4
    +(1-\alpha)[3(1-\alpha^2)+\alpha^3]z^3\notag\\
    &+(1-\alpha)[3+3\alpha(1-\alpha)+\alpha^4]z^2\notag\\
    &+[(1-\alpha)(1+2\alpha)+\alpha^5]z
      -\alpha^4(1-\alpha^2).
\end{align*}
This is Equation~\eqref{eq:phase-three-upper-polynomial}.
Every nonconstant coefficient is positive and the constant term is negative.
The polynomial is therefore strictly increasing on \(z>0\) and has exactly one positive root.
Substituting this root into Equation~\eqref{eq:proof-upper-parametrization} gives Equation~\eqref{eq:phase-three-upper-boundary}.

The parametrization also gives
\begin{equation}
  \frac1\alpha-\varphi
  =\frac{(1-\alpha)(1+\alpha+z)}
        {\alpha(1+z)[1+(1-\alpha)z]}.
\end{equation}
The right side decreases from \(1/\alpha-\alpha\) to zero, so \(z>0\) parametrizes the full interval \(\alpha<\varphi<1/\alpha\).
The linear term of the polynomial bounds its root by
\begin{equation}
  z_+
  <\frac{\alpha^4(1-\alpha^2)}
          {(1-\alpha)(1+2\alpha)+\alpha^5}
  <\alpha.
\end{equation}
On the other hand,
\begin{equation}
  \varphi=1
  \iff z(1+z)=\frac{\alpha}{1-\alpha},
\end{equation}
whose positive solution is larger than \(\alpha\).
Hence \(\alpha<\varphi_+<1\), and these equations have no further solution before recovery.
\end{proof}

\subsection{Long-time behavior}

Propositions~\ref{prop:lower-boundary} and \ref{prop:upper-boundary} determine the two boundaries.
Finally, we use the eigenvalues of the iteration matrix to obtain the large-\(t\) behavior.

\begin{proposition}[Long-time behavior]
\label{prop:long-time-cavity-phases}
Suppose \(0<\alpha<1\), \(\lambda\to0\), and \(\eta>0\) is sufficiently small.
Using the conditional averages in Proposition~\ref{prop:cavity-moments}, assume that the largest real part among the nonzero eigenvalues of \(((\mat X^{\mathrm{in}})^\top\mat X^{\mathrm{in}}/(\eta L))\mat A_c\) changes from negative to positive at \(\varphi_-\), changes from positive to negative at \(\varphi_+\), and does not change sign elsewhere.
Assume also that the normalized traces below have finite limits and that the inverse matrices in the bounded region have uniformly bounded normalized norms.
\begin{equation}
  \mathcal E(t)=
  \begin{cases}
    \Theta(t^2),
      &0<\varphi<\varphi_-(\alpha),\\
    \exp\!\bigl(\Theta(\eta t)\bigr),
      &\varphi_-(\alpha)<\varphi<\varphi_+(\alpha),\\
    O(1),
      &\varphi_+(\alpha)<\varphi<1/\alpha,\\
    \mathcal E_{\mathrm{GD}}(t),
      &\varphi>1/\alpha.
  \end{cases}.
  \label{eq:proof-long-time-cavity-phases}
\end{equation}
\end{proposition}

\begin{proof}
As \(\varphi\to(1/\alpha)^-\), \(u_\varphi,c_\varphi\to0\), \(v_\varphi\to1\), and \(\hat{\mat A}/\eta\to-\mat I_D\), so all nonzero eigenvalues have negative real part near \(1/\alpha\).
Together with Propositions~\ref{prop:lower-boundary} and
\ref{prop:upper-boundary}, the spectral assumption in the statement implies
that the nonzero eigenvalues have negative real part below \(\varphi_-\),
positive real part between \(\varphi_-\) and \(\varphi_+\), and negative real
part above \(\varphi_+\).  The discussion in
Subsection~\ref{subsec:spectrum-conditions} then gives
\(\mathcal E(t)=\Theta(t^2)\) for \(0<\varphi<\varphi_-\),
\(\mathcal E(t)=\exp(\Theta(\eta t))\) for
\(\varphi_-<\varphi<\varphi_+\), and \(\mathcal E(t)=O(1)\) for
\(\varphi_+<\varphi<1/\alpha\).  Equation~\eqref{eq:cavity-frobenius-error}
shows that the GD matrix has not yet been recovered in the last regime.

Finally, if \(\varphi>1/\alpha\), then \(\alpha\varphi>1\), and
Proposition~\ref{prop:rank-deficient-gd-recovery} gives
\(\mathcal E(t)=\mathcal E_{\mathrm{GD}}(t)\).
These arguments prove Equation~\eqref{eq:proof-long-time-cavity-phases}.
\end{proof}

Finally, \(0<\varphi_-<\alpha<\varphi_+<1<1/\alpha\) partitions the entire \(\alpha<1\) half-plane into the four open intervals in Result~\ref{res:long-time-phases}.
For \(\alpha>1\), Proposition~\ref{prop:gd-uniqueness} gives the exact recovery condition \(\varphi>1\), while Proposition~\ref{prop:projected-gd-long-time} gives the quadratic phase for \(0<\varphi<1\).
Together, these propositions give every region of Result~\ref{res:long-time-phases}.

\section{Experimental Details for Fully Trained Attention}
\label{app:generalization-audit}

This appendix provides the full protocol for the experiments in Section~\ref{sec:empirical-extensions} and Figure~\ref{fig:attention-nonmonotonic}.

\subsection{Data and architectures}
\label{app:attention-nonmonotonic}

We use the Gaussian regression task, teacher GD trajectories, prompt coordinates, and generalization error defined in Section~\ref{sec:model}.
The dimension is \(D=32\), the context ratio is \(\alpha=0.75\), the training CoT length is \(K=2\), the teacher step size is \(\eta=0.1\), and the observation-noise standard deviation is \(\sigma=0.1\).

Both architectures have total token width \(2D+2\).
The context-only mask in Section~\ref{sec:empirical-extensions} means that the
query is formed from the final state column,
\begin{align}
  \vb z_k
  &:={}(\mat E_{\mu,k})_{:,L+1}
  =\begin{pmatrix}\vb0_D\\0\\\vb s_{\mu,k}\\1\end{pmatrix},
\end{align}
whereas the keys and values are formed only from the first \(L\) context columns,
\begin{align}
  \begin{pmatrix}\vb c_1&\cdots&\vb c_L\end{pmatrix}
  &:={}(\mat E_{\mu,k})_{:,1:L}
  ={}
  \begin{pmatrix}
    \mat X_\mu\\
    \vb y_\mu^\top\\
    \mat0_{D\times L}\\
    \vb0_L^\top
  \end{pmatrix}
  \in\R^{(2D+2)\times L}.
\end{align}
For each head, the three bias-free projections are therefore
\begin{align}
  \vb q_k^h&={}\mat W_Q^h\vb z_k,
  &
  \vb k_\ell^h&={}\mat W_K^h\vb c_\ell,
  &
  \vb v_\ell^h&={}\mat W_V^h\vb c_\ell.
\end{align}
Only \(\vb z_k\) enters the query projection, and only
\(\vb c_1,\ldots,\vb c_L\) enter the key and value projections.  The resulting
head output is
\begin{align}
  \vb u_k^h
  &={}
  \frac{1}{L}\sum_{\ell=1}^{L}
  \bigl((\vb q_k^h)^\top\vb k_\ell^h\bigr)\vb v_\ell^h
  && \text{(linear attention)},
  \notag\\
  \vb u_k^h
  &={}
  \sum_{\ell=1}^{L}
  \frac{
    \exp\!\bigl((\vb q_k^h)^\top\vb k_\ell^h/\sqrt{d_h}\bigr)
  }{
    \sum_{j=1}^{L}
    \exp\!\bigl((\vb q_k^h)^\top\vb k_j^h/\sqrt{d_h}\bigr)
  }
  \vb v_\ell^h
  && \text{(softmax attention)}.
  \label{eq:context-only-attention}
\end{align}
Thus the state token is absent from the key/value sequence, rather than present
with a zero key and value.  

Linear attention has one head with \(d_1=2D+2\), learns the matrices \(\mat W_Q^1,\mat W_K^1,\mat W_V^1\in\R^{(2D+2)\times(2D+2)}\), and has no learned output projection.  
A fixed readout selects the \(D\) state coordinates of the head output, and the residual connection adds this increment to the current state:
\begin{equation}
  \hat{\vb s}_{\mu,k+1}
  =\vb s_{\mu,k}+(\vb u_k^1)_{D+2:2D+1}.
\end{equation}
Its Q/K/V weights use the default PyTorch linear-layer initialization, which samples each entry uniformly from
\([-1/\sqrt{2D+2},1/\sqrt{2D+2}]\), followed by multiplication by \(0.1\).

Softmax attention has two heads with \(d_1=d_2=D+1\).  It learns the bias-free
\(\mat W_Q^h,\mat W_K^h,\mat W_V^h\in\R^{(D+1)\times(2D+2)}\) and
\(\mat W_O\in\R^{(2D+2)\times(2D+2)}\).  The two head outputs are concatenated,
mapped by \(\mat W_O\), restricted to the same \(D\) state coordinates, and
added to the current state:
\begin{equation}
  \hat{\vb s}_{\mu,k+1}
  =\vb s_{\mu,k}
   +\left[
      \mat W_O
      \begin{pmatrix}\vb u_k^1\\ \vb u_k^2\end{pmatrix}
    \right]_{D+2:2D+1}.
\end{equation}

\subsection{Optimization and evaluation}

Each model is trained for 5,000 AdamW updates with initial learning rate \(0.003\), cosine decay to \(0.00003\), weight decay \(10^{-4}\), gradient-norm clipping at \(10\), and minibatches of at most 128 transitions.
For each seed, evaluation uses 192 independently generated test contexts and a recursive horizon of 128 steps.

% Temporarily omit the empirical mechanism-check details.
%\input{sections/empirical_details}

\end{document}